\documentclass[lettersize,journal]{IEEEtran}
\usepackage{amsmath,amsfonts}
\usepackage{array}
\usepackage[caption=false,font=normalsize,labelfont=sf,textfont=sf]{subfig}
\usepackage{textcomp}
\usepackage{stfloats}
\usepackage{url}
\usepackage{verbatim}
\usepackage{graphicx}
\def\BibTeX{{\rm B\kern-.05em{\sc i\kern-.025em b}\kern-.08em
    T\kern-.1667em\lower.7ex\hbox{E}\kern-.125emX}}
\usepackage{balance}

\usepackage{amssymb}
\usepackage{amsmath}
\usepackage{algorithm,algpseudocode}
\usepackage{algpseudocode}
\usepackage{xcolor}
\usepackage{graphicx}
\usepackage{tikz}
\usetikzlibrary{fit,calc}
\usepackage{booktabs}
\usepackage{multirow}

\usepackage{xr}
\makeatletter
\newcommand*{\addFileDependency}[1]{
  \typeout{(#1)}
  \@addtofilelist{#1}
  \IfFileExists{#1}{}{\typeout{No file #1.}}
}
\makeatother

\newcommand*{\myexternaldocument}[1]{
    \externaldocument{#1}
    \addFileDependency{#1.tex}
    \addFileDependency{#1.aux}
}

\myexternaldocument{supplementary}

\definecolor{omer_red}{HTML}{c00000}
\definecolor{omer_green}{HTML}{00b050}
\definecolor{omer_blue}{HTML}{2f5597}

\newcommand{\expnumber}[2]{{#1}\mathrm{e}{#2}}

\newcommand{\Review}[1]{{#1}}

\newcommand{\dR}{\mathbb{R}}
\newcommand{\dP}{\mathbb{P}}
\newcommand{\dN}{\mathbb{N}}
\newcommand{\dE}{\mathbb{E}}

\newcommand{\tP}{\mathbb{P}_{y|x}}
\newcommand{\thP}{\hat{\mathbb{P}}_{y|x}}

\newcounter{algsubstate}
\renewcommand{\thealgsubstate}{\alph{algsubstate}}

\newcommand*{\tikzmk}[1]{\tikz[remember picture,overlay,] \node[inner sep=0mm,outer sep=0mm] (#1) {};\ignorespaces}
\newcommand{\boxit}[1]{\tikz[remember picture,overlay]{\node[inner sep=0mm,outer sep=0mm,xshift=-36pt,fill=#1,opacity=.25,fit={(A)($(B)+(1.10\linewidth,-0.1\baselineskip)$)}] {};}\ignorespaces}
\colorlet{pink}{red!40}
\colorlet{blue2}{cyan!60}
\colorlet{green}{green!40}

\begin{document}
\title{Principal Uncertainty Quantification with Spatial Correlation for Image Restoration Problems}
\author{Omer Belhasin, Yaniv Romano, Daniel Freedman, Ehud Rivlin, Michael Elad
\thanks{O. Belhasin (omerbe@verily.com), D. Freedman (danielfreedman@verily.com), Ehud Rivlin (ehud@verily.com) and M. Elad (melad@verily.com) are with Verily Life Sciences, Israel.}
\thanks{O. Belhasin (omer.be@cs.technion.ac.il) and Y. Romano (yromano@technion.ac.il) are with the Department of Computer Science, Technion - Israel Institute of Technology, Haifa, Israel.}
}

\markboth{IEEE Transactions on Pattern Analysis and Machine Intelligence}
{Omer Belhasin, \MakeLowercase{\textit{(et al.)}: Principal Uncertainty Quantification with Spatial Correlation for Image Restoration Problems}}

\maketitle

\begin{abstract}
Uncertainty quantification for inverse problems in imaging has drawn much attention lately.
Existing approaches towards this task define uncertainty regions based on probable values per pixel, while ignoring spatial correlations within the image, resulting in an exaggerated volume of uncertainty.
In this paper, we propose PUQ (Principal Uncertainty Quantification) -- a novel definition and corresponding analysis of uncertainty regions that takes into account spatial relationships within the image, thus providing reduced volume regions.
Using recent advancements in generative models, we derive uncertainty intervals around principal components of the empirical posterior distribution, forming an ambiguity region that guarantees the inclusion of true unseen values with a user-defined confidence probability.
To improve computational efficiency and interpretability, we also guarantee the recovery of true unseen values using only a few principal directions, resulting in more informative uncertainty regions.
Our approach is verified through experiments on image colorization, super-resolution, and inpainting; its effectiveness is shown through comparison to baseline methods, demonstrating significantly tighter uncertainty regions.
\end{abstract}


\section{Introduction}

\IEEEPARstart{R}{estoration} tasks in imaging are widely encountered in various disciplines, including cellular cameras, surveillance, experimental physics,  and medical imaging. These inverse problems are broadly defined as the need to recover an unknown image given corrupted measurements of it. Such problems, e.g., colorization, super-resolution, and inpainting, are typically ill-posed, implying that multiple solutions can explain the unknown target image.
In this context, uncertainty quantification aims to characterize the range of possible solutions, their spread, and variability.
This has an especially important role in applications such as astronomy and 
medical diagnosis, where it is necessary to establish statistical boundaries for possible gray-value deviations.
The ability to characterize the range of permissible solutions with accompanying statistical guarantees has thus become an important and useful challenge, addressed in this paper.

Prior work on this topic~\cite{angelopoulos2022image,horwitz2022conffusion} has addressed the uncertainty assessment by constructing intervals of possible values for each pixel via quantile regression \cite{koenker1978regression}, or other heuristics such as estimations of per-pixel residuals.
While this line of thinking is appealing due to its simplicity, it disregards spatial correlations within the image, and thus provides an exaggerated uncertainty range. 
The study in \cite{sankaranarayanan2022semantic} has improved the above by quantifying the uncertainty in a latent space, thus taking spatial dependencies into account. However, by relying on a non-linear, non-invertible and uncertainty-oblivious transformation, this method suffers from interpretability limitations -- See Section~\ref{sec: related work} for further discussion.

\begin{figure}[t]
    \centering
    \includegraphics[width=0.52\textwidth,trim={1cm 0cm 0cm 0cm},clip]{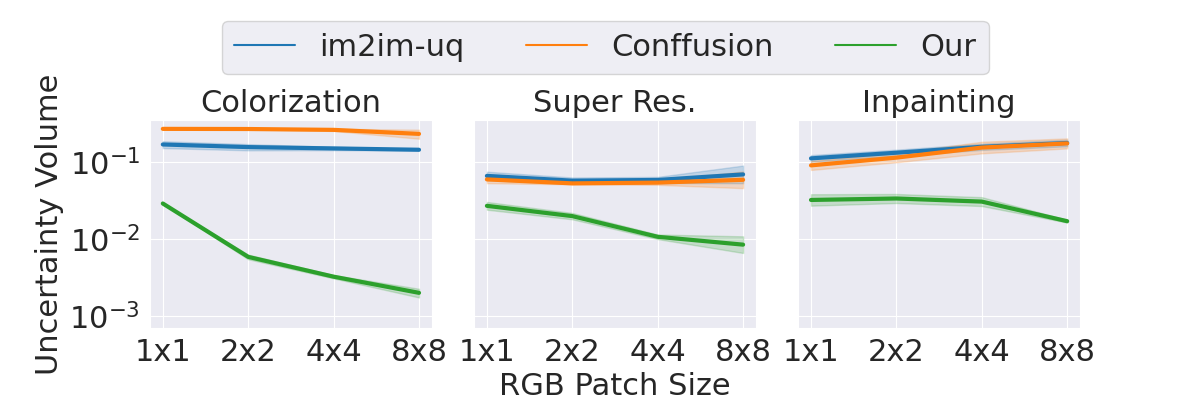}
    \vspace{-0.3cm}
    \caption{Comparison of PUQ's performance on the CelebA-HQ dataset in image colorization, super-resolution, and inpainting tasks using the E-PUQ procedure (Section~\ref{sec: Exact PUQ}) applied on RGB image patches of varying size. As seen, our method provides tighter uncertainty regions with significantly smaller uncertainty volumes ($ \times10$ in super-res. and inpainting, and $ \times 100 $ in colorization). The compared methods are im2im-uq~\cite{angelopoulos2022image} and Conffusion~\cite{horwitz2022conffusion}.}
    \label{fig: main}
\end{figure}

In this paper, we propose \emph{Principal Uncertainty Quantification} (PUQ) -- a novel approach that accounts for spatial relationships while operating in the image domain, thus enabling a full and clear interpretation of the quantified uncertainty region. PUQ uses the principal components of the empirical posterior probability density function, which describe the spread of possible solutions.
PCA essentially approximates this posterior by a Gaussian distribution that tightly encapsulates it. Thus,
this approach reduces the uncertainty volume\footnote{The definition of this volume, which plays a critical part in this work, is further discussed in later sections and given in Equation~\eqref{eq: uncertainty volume}.}, as demonstrated in Figure~\ref{fig: main}.
This figure presents a comparison between our proposed Exact PUQ procedure (see Section~\ref{sec: Exact PUQ}) and previous work~\cite{angelopoulos2022image,horwitz2022conffusion}, showing a much desired trend of reduced \emph{uncertainty volume} that further decreases as the size of the patch under consideration grows. 


Our work aims to improve the quantification of the uncertainty volume 
by leveraging recent advancements in generative models serving as stochastic solvers for inverse problems. While our proposed approach is applicable using any such solver (e.g., conditional GAN~\cite{mirza2014conditional}), we focus in this work on 
diffusion-based techniques, which have recently emerged as the leading image synthesis approach, surpassing GANs and other alternative generators \cite{dhariwal2021diffusion}.
Diffusion models offer a systematic and well-motivated algorithmic path towards the task of sampling from a prior probability density function (PDF), $\dP_y$, through the repeated application of a trained image-denoiser \cite{sohl2015deep, ho2020denoising}.
An important extension of these models allows the sampler to become conditional, drawing samples from the posterior PDF, $\tP$, where $x$ represents the observed measurements.
This approach has recently gained significant attention \cite{dhariwal2021diffusion, ho2022cascaded, saharia2022image, saharia2022palette}, yielding a  fascinating viewpoint to inverse problems, in which a variety of candidate high perceptual quality solutions to such problems are obtained.

\begin{figure}[t]
    \centering
    \includegraphics[width=0.5\textwidth,trim={6cm 3cm 2cm 3cm},clip]{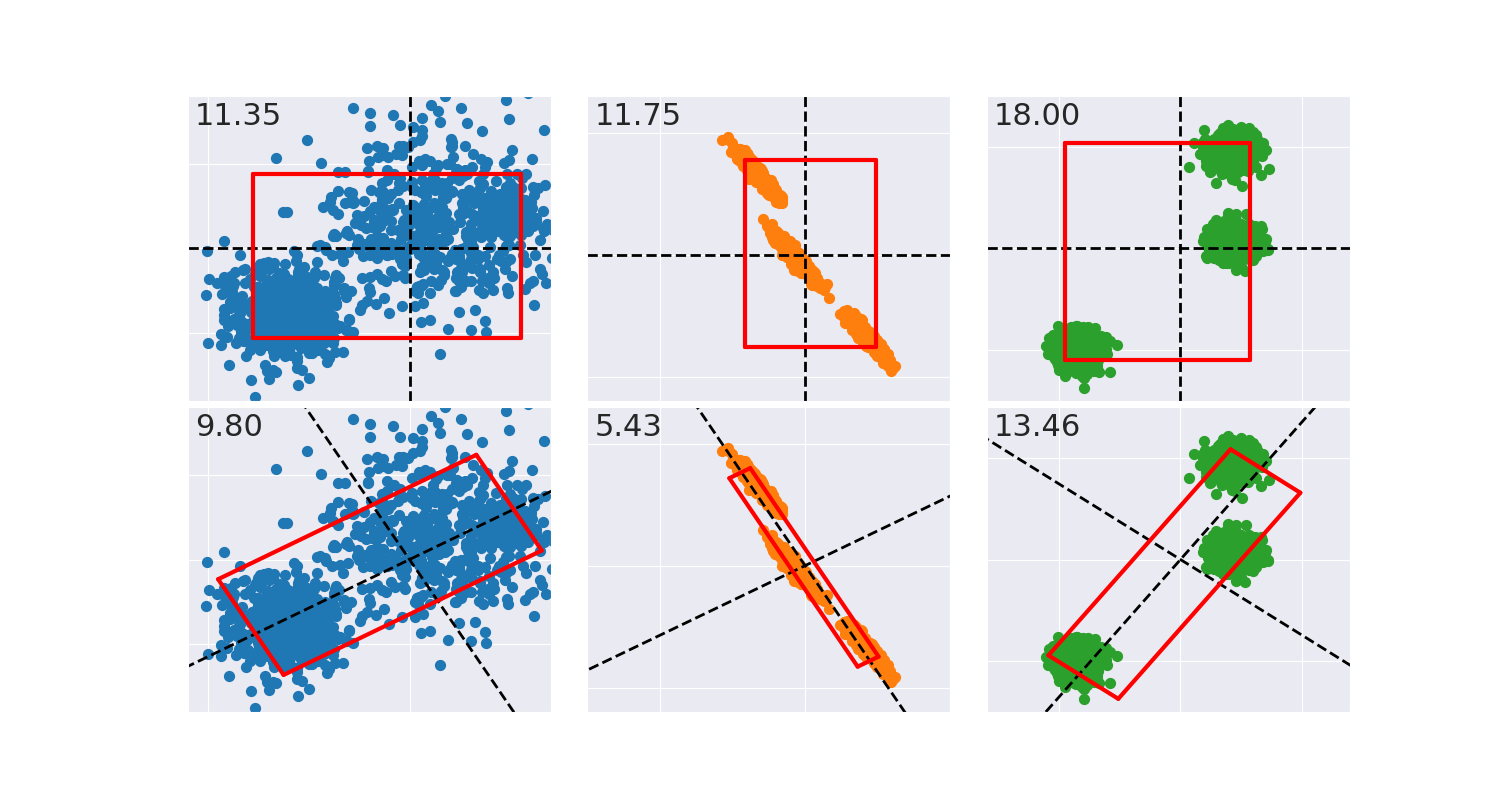}
    \vspace{-0.3cm}
    \caption{An illustration of uncertainty regions (in red) of 2d posterior distributions and considering three different PDF behaviors, shown in blue, orange, and green.
    The uncertainty regions are formed from intervals, as defined in Equation~\eqref{eq: intervals}, where $\hat{l}(x)$ and $\hat{u}(x)$ represent the $0.05$ and $0.95$ quantiles over the dashed black axes.
    The top row presents the uncertainty region in the pixel domain using standard basis vectors that ignores the spatial correlations, while the lower row presents the regions using the principal components as the basis. The uncertainty volume, defined in Equation~\eqref{eq: uncertainty volume}, is indicated in the top left corner of each plot. The $90\%$ coverage guarantee, outlined in Equation~\eqref{eq: theoretical coverage gaurentee} with $w_i := 1/2$, is satisfied by all. As can be seen, the lower row regions take spatial dependencies into account and are significantly smaller than the pixelwise corresponding regions in the upper row.}
    \label{fig: 2d example}
\end{figure}

In this work, we generalize the pixelwise uncertainty assessment, as developed in~\cite{angelopoulos2022image,horwitz2022conffusion}, so as to incorporate spatial correlations between pixels.  
This generalization is obtained by considering an image-adaptive basis for a linear space that replaces the standard basis in the pixelwise approach.
To optimize the volume of the output uncertainty region, we propose a statistical analysis of the posterior obtained from a diffusion-based sampler (e.g., \cite{saharia2022image,saharia2022palette}), considering a series of candidate restorations.
Our method may be applied both globally (on the entire image) or locally (on selected portions or patches), yielding a tighter and more accurate encapsulation of statistically valid uncertainty regions.
For the purpose of adapting the basis, we compute and leverage the principal components of the candidate restorations.
As illustrated in Figure~\ref{fig: 2d example} for a simple 2-dimensional PDFs, the pixelwise regions are less efficient and may contain vast empty areas, and  especially so in cases where pixels exhibit strong correlation. 
Clearly, as the dimension increases, the gap between the standard and the adapted uncertainty quantifications is further amplified.

Our proposed method offers two \emph{conformal prediction} \cite{vovk2005algorithmic, lei2018distribution, angelopoulos2021gentle} based calibration options (specifically, using the Learn then Test \cite{angelopoulos2021learn} scheme) for users to choose from, with a trade-off between precision and complexity.
These include (i) using the entire set of principal components, (ii) using a predetermined subset of them\footnote{We also propose a reduced complexity variation of this option that controls the number of necessary principal components to be used.}.
The proposed calibration procedures ensure the validity of the uncertainty region to contain the unknown true values with a user-specified confidence probability, while also ensuring the recovery of the unknown true values using the selected principal components when only a subset is used.
Applying these approaches allows for efficient navigation within the uncertainty region of highly probable solutions.

We conduct various local and global experiments to verify our method, considering three challenging tasks: image colorization, super-resolution, and inpainting, all described in Section~\ref{sec: results}, and all demonstrating the advantages of the proposed approach. For example, when applied locally on $8$$\times$$8\times$$3$ patches, our experiments show a reduction in the guaranteed uncertainty volume by a factor of $\sim$$10$-$100$ compared to previous approaches, as demonstrated in Figure~\ref{fig: main}. Moreover, this local approach can have a substantially reduced computational complexity while retaining the statistical guarantees, by drawing far fewer posterior samples and using a small subset of the principal components. As another example, the global tests on the colorization task provide an unprecedented tightness in uncertainty volumes. This is accessible via a reduced set of drawn samples, while also allowing for efficient navigation within the solution set. 

\noindent In summary, our contributions are the following:
\begin{enumerate}
\item We introduce a novel generalized definition of uncertainty region that  leverages an adapted linear-space basis for better posterior coverage.
\item We propose a new method for quantifying the uncertainty of inverse problems that considers spatial correlation, thus providing tight uncertainty regions.
\item We present two novel calibration procedures for the uncertainty quantification that provide statistical guarantees for unknown data to be included in the uncertainty region with a desired coverage ratio while being recovered with a small error by the selected linear axes.
\item We provide a comprehensive empirical study of three challenging image-to-image translation tasks:  colorization, super-resolution, and inpainting, demonstrating the effectiveness of the proposed approach in all modes.
\end{enumerate}

\section{Related Work}
\label{sec: related work}

\noindent Inverse problems in imaging have been extensively studied 
over the years; this domain has been deeply influenced by the AI revolution~\cite{lathuiliere2019comprehensive, santhanam2017generalized, yan2016attribute2image, gregor2015draw, mirza2014conditional}.
A promising recent approach towards image-to-image translation problems relies on the massive progress made on learned generative techniques. These new tools enable to model the conditional distribution of the output images given the input, 
offering a fair sampling from this PDF. 
Generative-based solvers of this sort create a new and exciting opportunity for getting high perceptual quality solutions for the problem in hand, while also accessing a diverse set of such candidate solutions. 

Recently, \emph{Denoising Diffusion Probabilistic Models} (DDPM) \cite{sohl2015deep,ho2020denoising} have emerged as a new paradigm for image generation, surpassing the state-of-the-art results achieved by GANs \cite{goodfellow2020generative,dhariwal2021diffusion}.
Consequently, several \emph{conditional} diffusion methods have been explored \cite{dhariwal2021diffusion,ho2022cascaded, saharia2022image,saharia2022palette}, including SR3 \cite{saharia2022image} -- a diffusion-based method for image super-resolution, Palette \cite{saharia2022palette} -- a diffusion-based unified framework for image-to-image translation tasks, and more (e.g.~\cite{song2019generative,kadkhodaie2020solving,song2020score,kawar2021stochastic,kawar2021snips,kawar2022denoising,kadkhodaie2023learning}). 
Note that current conditional algorithms for inverse problems do not offer statistical guarantees  against model deviations and hallucinations.

Moving to uncertainty quantification, the field of machine learning has been seeing rich work on the derivation of statistically rigorous confidence intervals for predictions \cite{romano2019conformalized,chernozhukov2021distributional,sesia2021conformal,gupta2022nested,kivaranovic2020adaptive}.
One key paradigm in this context is \emph{conformal prediction} (CP) \cite{vovk2005algorithmic, lei2018distribution, angelopoulos2021gentle} and \emph{risk-controlling} methods \cite{bates2021distribution,angelopoulos2021learn,angelopoulos2022conformal}, which allow to rigorously quantify the prediction uncertainty of a machine learning model with a user-specified probability guarantee.
Despite many proposed methods, only a few have focused on mitigating uncertainty assessment in image restoration problems, including im2im-uq \cite{angelopoulos2022image} and Conffusion \cite{horwitz2022conffusion}.
The work reported in~\cite{teneggi2023trust} is closely related as it introduced a generalized, and thus improved, calibration scheme for Conffusion \cite{horwitz2022conffusion}.
All these works have employed a risk-controlling paradigm \cite{
bates2021distribution} to provide statistically valid prediction intervals over the pixel domain, ensuring the inclusion of ground-truth solutions in the output intervals.
However, these approaches share the same limitation of operating in the pixel domain while disregarding spatial correlations within the image or the color layers. This leads to an unnecessarily exaggerated volume of uncertainty. 

An exception to the above is~\cite{sankaranarayanan2022semantic}, which quantifies uncertainty in the latent space of GANs. Their migration from the image domain to the latent space is a rigid, global, non-linear, non-invertible and uncertainty-oblivious transformation. Therefore, quantification of the uncertainty in this domain is quite limited. More specifically, rigidity implies that this approach cannot adapt to the complexity of the problem by adjusting the latent space dimension; Globality suggests that it cannot be operated locally on patches in order to better localize the uncertainty assessments; Being non-linear implies that an evaluation of the uncertainty volume (see Section~\ref{sec: Problem Formulation}) in the image domain is hard and next to impossible; Non-invertability of means that some energy is lost from the image in the analysis and not accounted for, thus hampering the validity of the statistical guarantees; Finally, note that the latent space is associated with the image content, but does not represent the prime axes of the uncertainty behavior. 
Note that due to the above, and especially the inability to provide certified volumes of uncertainty, an experimental comparison of our method to~\cite{sankaranarayanan2022semantic} is impossible. 

Inspired by the above contributions, we propose a novel alternative uncertainty quantification approach that takes spatial relationships into account. Our work provides tight uncertainty regions, compared to prior work, with user-defined statistical guarantees through the use of a CP-based paradigm. Specifically, we adopted the Learn then Test \cite{angelopoulos2021learn} that provides statistical guarantees for controlling multiple risks.


\section{Problem Formulation}
\label{sec: Problem Formulation}

\noindent Let $\dP_{x,y}$ be a probability distribution over $\mathcal{X} \times \mathcal{Y}$, where $\mathcal{X}$ and $\mathcal{Y}$ represent the input and the output space, respectively, for the inverse problem at hand. E.g., for the task of image colorization, $\mathcal{Y}$ could represent full-color high-quality images, while $\mathcal{X}$ represents their colorless versions to operate on. We assume that $\mathcal{X,Y} \subset [0,1]^d \subset \dR^d$, where, without loss of generality, $d$ is assumed to be the dimension of both spaces.
In the context of examining patches within output images, we define $\mathcal{Y}_{\text{patch}}$ as the patch space of the output images. For simplicity, we use the same notation, $d$, for $\mathcal{Y}$ and $\mathcal{Y}_{\text{patch}}$, while it is clear that the dimension of $\mathcal{Y}_{\text{patch}}$ is smaller and controlled by the user through the patch size to work on.

Given an input measurement $x \in \dR^d$, we aim to quantify the uncertainty of the possible solutions to the inverse problem, as manifested by the estimated $d$-dimensional posterior distribution, $\thP$. 
The idea is to enhance the definition of pixelwise uncertainty intervals 
by integrating the spatial correlations between pixels to yield a better structured uncertainty region. To achieve this, we propose to construct uncertainty intervals using a designated collection of orthonormal basis vectors for $\dR^d$ instead of intervals over individual pixels. We denote this collection by $\hat{B}(x) = \{ \hat{v}_1(x), \hat{v}_2(x) \dots \hat{v}_d(x) \}$, where $\hat{v}_i(x) \in \dR^d$. These vectors are instance-dependent, thus best adapted to their task. 
An intuitive example of such a basis is the standard one, $\hat{B}(x) = \{e_1, e_2 \dots e_d\}$, where $e_i\in\dR^d$ is the one-hot vector with value 1 in the $i^{th}$ entry. 
In our work, we use a set of principal components of $\thP$, which will be discussed in detail in Section~\ref{sec: method}.

Similar to \cite{angelopoulos2022image, horwitz2022conffusion}, we use an interval-based method centered around the conditional mean image, i.e., an estimate of $\dE[y|x] \in \dR^d$, denoted by $\hat{\mu}(x)$.
Formally, we utilize the following interval-valued function that constructs prediction intervals along each basis vector around the estimated conditional mean:
{\small\begin{align}
\label{eq: intervals}
\mathcal{T}(x; \hat{B}(x))_i := \Big[ {\hat{v}_i(x)}^T \hat{\mu}(x) - \hat{l}(x)_i , {\hat{v}_i(x)}^T \hat{\mu}(x) + \hat{u}(x)_i \Big] .
\end{align}}
In the above, $i\in \{1,2 \dots d\}$ is a basis vector index, and $\hat{l}(x)_i \in \dR^+$ and $\hat{u}(x)_i \in \dR^+$ are the lower and upper interval boundaries for the projected values of candidate solutions emerging from $\thP$. That is, if $\hat{y} \sim \thP$ is such a solution, ${\hat{v}_i(x)}^T \hat{y}$ is its $i$-th projection, and this value should fall within 
$\mathcal{T}(x; \hat{B}(x))_i$ with high probability. 
Returning to the example of the standard basis, the above equation is nothing but  pixelwise prediction intervals, which is precisely the approach taken in \cite{angelopoulos2022image, horwitz2022conffusion}.
By leveraging this generalization, the uncertainty intervals using these basis vectors form a $d$-dimensional hyper-rectangle,  
referred to as the \emph{uncertainty region}.

Importantly, we propose that the interval-valued function, $\mathcal{T}$, should produce valid intervals that contain a user-specified fraction of the projected ground-truth values within a risk level of $\alpha \in (0,1)$. In other words, more than $1-\alpha$ of the projected ground-truth values should be contained within the intervals, similar to the approach taken in previous work in the pixel domain.
To achieve this, we propose a holistic expression that aggregates the effect of all the intervals, 
$\mathcal{T}(x; \hat{B}(x))$. This expression leads to the following condition:
{\small\begin{equation}
\label{eq: theoretical coverage gaurentee}
\dE \left[ \sum_{i=1}^d \hat{w}_i(x) \cdot \mathbf{1}\left\{\hat{v}_i(x)^T y \in \mathcal{T}(x; B(x))_i \right\} \right] > 1 - \alpha,
\end{equation}} 
where $y \in \dR^d$ is the unknown ground-truth and $\hat{w}_i(x)\in[0,1]$ s.t. $\sum_{i=1}^d \hat{w}_i(x) = 1$ are the weight factors that set the importance of covering the projected ground-truth values along each interval.
In Section~\ref{sec: method} we discuss the proposed holistic expression and a specific choice of these weights.
As an example, we could set $\alpha=0.1$ and $\hat{w}_i(x) := 1/d$, indicating that more than $90\%$ of the projected ground-truth values onto the basis vectors are contained in the intervals, as illustrated in a 2d example in Figure~\ref{fig: 2d example} for different kinds of $\thP$.

As discussed above and demonstrated in Figure \ref{fig: 2d example}, if the orthonormal basis in Equation~\eqref{eq: intervals} is chosen to be the standard one, we get the pixel-based intervals
that disregard spatial correlations within the image, thus leading to an exaggerated uncertainty region. 
In this work, we address this limitation by transitioning to an instance-adapted orthonormal basis of $\dR^d$ that allows the description of uncertainty using axes that are not necessarily pixel-independent, thereby providing tighter uncertainty regions.
While such a basis could have been defined analytically using, for example, orthonormal wavelets \cite{meyer1990orthonormal},
we suggest a learned and thus a better-tuned one.
The choice to use a linear and orthonormal representation for the uncertainty quantification comes as a natural extension of the pixelwise approach, retaining much of the simplicity and efficiency of treating each axis separately. Note that the orthogonality enables the decomposition of $y$ around $\hat{\mu}(x)$ via its projected values, $y = \hat{\mu}(x) + \sum_{i=1}^d \left[\hat{v}_i(x)^T (y-\hat{\mu}(x))\right]  \hat{v}_i(x)$, which we refer to as the \emph{exact reconstruction property}.



To evaluate the uncertainty across different uncertainty regions, we introduce a new metric called the \emph{uncertainty volume}, $\mathcal{V}(x;\mathcal{T}(x;\hat{B}(x)))$, which represents the $d^{th}$ root of the uncertainty volume with respect to intervals $\mathcal{T}(x;\hat{B}(x))$, defined in the following equation: 
{\small\begin{align}
\label{eq: uncertainty volume}
\mathcal{V}(x;\mathcal{T}(x;&\hat{B}(x))) := \sqrt[\uproot{10}d]{\prod_{i=1}^d \left[ \hat{u}(x)_i + \hat{l}(x)_i \right]} \\
\nonumber&\approx \exp \left( \frac{1}{d} \sum_{i=1}^d \log \left( \hat{u}(x)_i + \hat{l}(x)_i + \epsilon \right) \right)-\epsilon  ~,
\end{align}}
where $\epsilon > 0$ is a small hyperparameter used for numerical stability.
In Section~\ref{sec: results} we demonstrate that our approach results in a significantly reduction in these uncertainty volumes when compared to previous methods.

When operating in high dimensions (e.g. on the full image), 
providing uncertainty intervals for all the $d$-dimensions poses severe challenges, both in complexity and interpretability.
In this case, 
constructing and maintaining the basis vectors becomes infeasible. Moreover, the uncertainty quantification using these intervals may be less intuitive compared to the conventional pixelwise approach because of the pixel-dependency between the basis vectors, which makes it difficult to communicate the uncertainty to the user.
To mitigate these challenges, we propose 
an option of using $K\ll d$ 
basis vectors that capture the essence of the uncertainty. 
In Section~\ref{sec: method}, we discuss how to dynamically adjust $K$ to provide fewer axes. 

While reducing the number of basis vectors benefits in interpretability and complexity, 
this option does not fulfill the exact reconstruction property. Therefore, we propose an extension to the conventional coverage validity of Equation~\eqref{eq: theoretical coverage gaurentee} that takes into account 
the reconstruction error of the decomposed ground-truth images. 
Specifically, the user sets a ratio of pixels, $q \in \dR$, and a maximum acceptable reconstruction error
over this ratio, $\beta\in (0,1)$.
This approximation allows us to reduce the number of basis vectors used to formulate $\hat{B}(x)$, such that the reconstruction will be valid according to the following condition: 
{\small\begin{align}
    \label{eq: theoretical reconstruction gaurentee}
    \dE \Biggl[ \hat{\mathcal{Q}}_q \Biggl( \left\{ \left| \sum_{j=1}^K \hat{v}_j(x)^T y_c \hat{v}_j(x) - y_c \right|_i \right\}_{i=1}^d \Biggr) \Biggr] \leq \beta ~,
\end{align}}
where $y_c:= y-{\hat \mu}(x)$ is the ground-truth image centered around ${\hat \mu}(x)$, and $\hat{\mathcal{Q}}_q(\cdot)$ is the empirical quantile function defined by the smallest $z$ satisfying $\frac{1}{d} \sum_{i=1}^d \mathbf{1} \{ z_i \leq \hat{\mathcal{Q}}_q(z) \} \geq q$.
In Section~\ref{sec: method}, we discuss this expression for assessing the validity of the basis vectors.
As an example, setting $q=0.9$ and $\beta=0.05$ would mean that the maximal reconstruction error of $90\%$ of the ground-truth pixels is no more than $5\%$ of the $[0,1]$ dynamic range.

\begin{figure}[t]
    \centering
    {\hspace{-1cm}\small $y$ ~~~~~~~~~~~~ $x$ ~~~~~~~~~~~~~~~~~~~~~~~~~ $\hat{y}_i \sim \thP$}
    \includegraphics[width=\columnwidth,trim={1cm 1cm 1cm 1cm},clip]{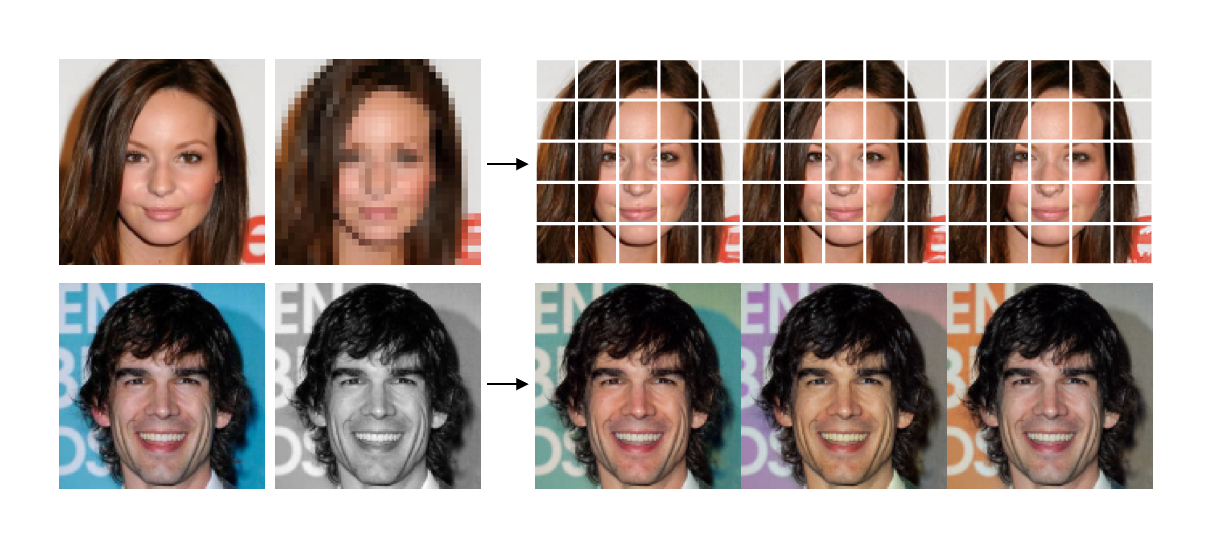}
    \caption{The sampling procedure for two image restoration problems using a conditional stochastic generator. The top row corresponds to super-resolution in local mode with patches, while the bottom row shows colorization in global mode. The implementation details are described in Section~\ref{sec: our implementation}.}
    \label{fig: sampling}
\end{figure}

\begin{figure}[t]
    \centering
    {\hspace*{0.05cm}\small Pixels Axes ~~~~~ PCs Axes ~~~~  Approximation ~~~ Calibration}
    \includegraphics[width=\columnwidth,trim={1cm 1cm 1cm 1.3cm},clip]{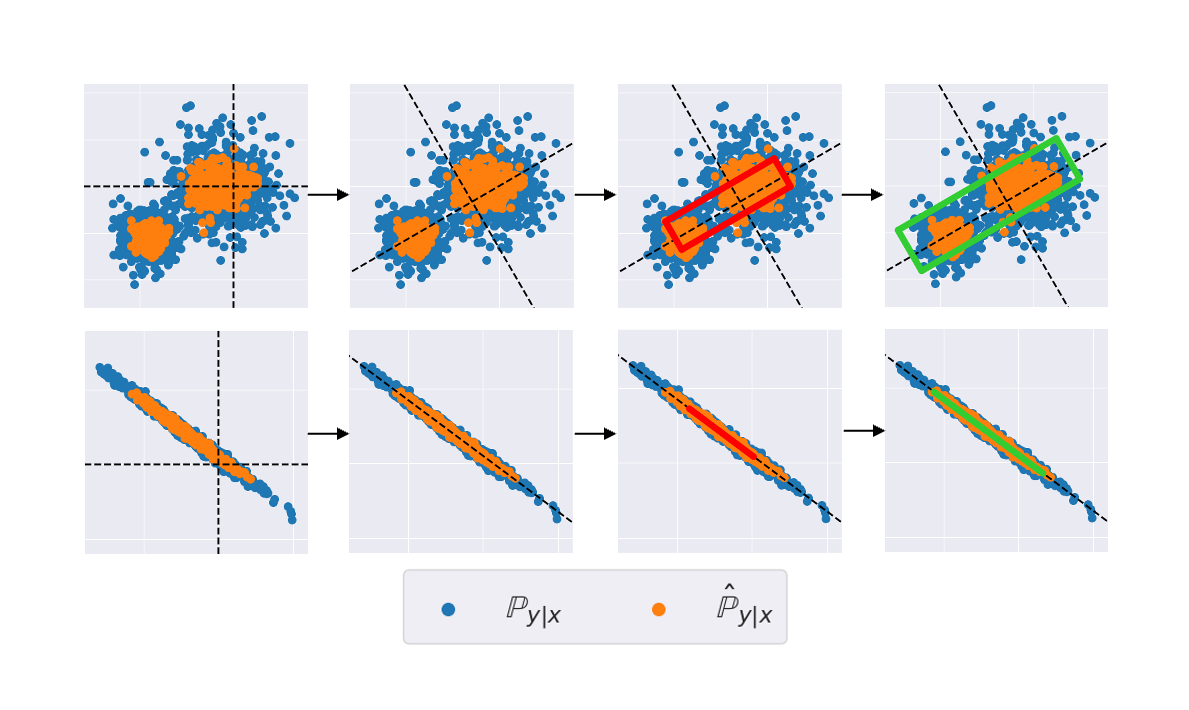}
    \caption{Illustration of our PUQ procedure in 2D ($d=2$) for a single instance $x\in\mathcal{X}$. The top row corresponds to the case when $K=d=2$ (as in E-PUQ), while the bottom row depicts the case when $K=1<d=2$ (as in DA-PUQ and RDA-PUQ). The procedure begins by drawing samples $\hat{y}_i\sim\thP$. Next, these samples are projected onto the PCs domain: $\hat{V}^T\hat{y}_i$, where $\hat{V} := [\hat{v}_1,\dots,\hat{v}_K]\in\dR^{d \times K}$. Then, we compute bounds along the PCs to contain the samples at the correct ratio, forming the intervals specified in Equation~\eqref{eq: intervals}. Finally, the intervals are scaled to statistically guarantee Equation~\eqref{eq: theoretical coverage gaurentee} and contain the correct ratio over solutions for unseen input instances. In the bottom row, the procedure also statistically guarantees Equation~\eqref{eq: theoretical reconstruction gaurentee} by ensuring a small recovery error of solutions to unseen input instances, as demonstrated by the small variance around the single PC.}
    \label{fig: calibration}
\end{figure}


\section{PUQ: Principal Uncertainty Quantification}
\label{sec: method}

\noindent In this section, we present \emph{Principal Uncertainty Quantification} (PUQ), our method for quantifying the uncertainty in inverse problems while taking into account spatial dependencies between pixels. PUQ uses the \emph{principal components} (PCs) of the solutions to the inverse problem for achieving its goal.
In Appendix~\ref{app: pcs motivation}, we provide an intuition behind the choice of the PCs as the basis.
Our approach can be used either globally across the entire image, referred to as the \emph{global mode}; or locally within predefined patches or segments of interest, referred to as the \emph{local mode}.
Local uncertainty quantification can be applied to any task, where the dimensionality of the target space is fully controlled by the user.
In contrast, global quantification is particularly advantageous for tasks that exhibit strong spatial correlations between pixels.

Our proposed method consists of two phases. In the first, referred to as the \emph{approximation phase}, a machine learning system is trained to predict the PCs of possible solutions, denoted by $\hat{B}(x) = \{ \hat{v}_1(x), \hat{v}_2(x), \dots, \hat{v}_K(x) \}$ (where $K \leq d$), as well as a set of importance weights, $\hat{w}(x) \in \dR^K$, referring to the vectors in $\hat{B}(x)$. In addition, the system estimates the necessary terms in Equation~\eqref{eq: intervals}, which include the conditional mean, $\hat{\mu}(x) \in \dR^d$, and the lower and upper bounds, $\tilde{l}(x) \in \dR^K$ and $\tilde{u}(x) \in \dR^K$ \footnote{Note that these bounds are meant for $\thP$ and not for $\tP$, and thus marked with tilde. The $\hat{l}(x), \hat{u}(x)$ bounds that are related to Equation~\eqref{eq: intervals} are defined later in the calibration schemes (Section~\ref{sec: calibration}).}, for the spread of projected solutions over $\hat{B}(x)$. 
All these ingredients are obtained by a diffusion-based conditional sampler as described in Figure~\ref{fig: sampling}. More details on this computational process are brought 
in Section~\ref{sec: our implementation}.

The above-described approximation phase is merely an estimation, as the corresponding heuristic intervals of Equation~\eqref{eq: intervals} may not contain the projected ground-truth values with a desired ratio.
Additionally, the basis vectors may not be able to recover the ground-truth pixel values within an acceptable threshold when $K < d$, or the basis set may contain insignificant axes in terms of variability.
Therefore, in the second, \emph{calibration phase}, we offer two calibration procedures on an held-out set of calibration data, denoted by $\mathcal{S}_\text{cal} := \{ ( x_i , y_i ) \}_{i=1}^n$. These assess the validity of our proposed uncertainty region over unseen data, which is composed by the intervals defined in Equation~\eqref{eq: intervals}. The choice between the two calibration procedures depends on the user,  taking into account the trade-off between precision and complexity. The steps of our proposed method are summarized in Algorithm~\ref{alg: general scheme}, and the two calibration strategies are as follows:

\noindent \textbf{(1) Exact PUQ} (E-PUQ - Section~\ref{sec: Exact PUQ}): In the setting of an exact uncertainty assessment, while assuming that $d$ PCs can be constructed and maintained in full, the exact reconstruction property is satisfied. Consequently, the calibration procedure is straightforward, involving only scaling of the intervals until they contain the 
user-specified miscoverage preference, denoted by $\alpha\in(0,1)$, of the projected ground-truth values falling outside the uncertainty region. This is similar to the approach taken in
previous work over the pixel domain.

\noindent \textbf{(2) Dimension-Adaptive PUQ} (DA-PUQ - Section~\ref{sec: Dimensional Adaptive PUQ}, RDA-PUQ - Appendix~\ref{app: Reduced Dimensional Adaptive PUQ}): In the setting of an approximate uncertainty assessment, while allowing for a small recovery error of projected ground-truth instances to full-dimensional instances, either due to complexity or interpretability reasons (see Section~\ref{sec: Problem Formulation}), the exact reconstruction property is no longer satisfied. Hence, in addition to the scaling procedure outlined above, we must verify that the $K$ PCs can decompose the ground-truth pixel values with a small error. In this calibration process, we also control the minimum number of the first $\hat{k}(x)$ 
PCs out of the $K$ PCs, such that a small reconstruction error can be guaranteed for unseen data. This number is dynamically determined per input image, so that instances with greater pixel correlations are assigned more PCs than those with weaker correlations.
As manually determining $K$ might be challenging, we introduce the Reduced Dimension-Adaptive PUQ (RDA-PUQ) procedure that  also controls that value as part of the calibration - see Appendix~\ref{app: Reduced Dimensional Adaptive PUQ}.


\begin{algorithm}[htb]
\small
\caption{Generating PUQ Axes and Intervals}
\label{alg: general scheme}
\begin{algorithmic}[1]

\Require{Training set. Calibration set. Number of PCs $K \in \dN$. An unseen input instance $x \in \dR^d$.}

\Ensure{Statistically valid uncertainty axes and intervals for $x$.}

\vspace{-5pt}
\tikzmk{A} \Comment{Approximation phase}

\State 
Train a machine learning system (e.g., Section~\ref{sec: our implementation}) to estimate the following:

    $K$ 
    PCs of $\thP$

    Importance weights of PCs
    
    The conditional mean
    
    Lower and upper bounds on the PCs

\vspace{-5pt}\tikzmk{B}\boxit{pink}

\vspace{-10pt}
\tikzmk{A} \Comment{Calibration phase}


\If{Exact uncertainty (accurate)}
    \State Apply E-PUQ using the calibration data
\ElsIf{Approximate uncertainty (reduced complexity)}
    \State Apply DA-PUQ / RDA-PUQ using the calibration data
\EndIf

\vspace{-5pt}\tikzmk{B}\boxit{blue2}

\vspace{-10pt}
\tikzmk{A} \Comment{Inference}

\State Provide statistically valid
uncertainty axes and intervals in terms of Equation~\eqref{eq: theoretical coverage gaurentee} and Equation~\eqref{eq: theoretical reconstruction gaurentee}, applied to an unseen input instance $x$

\vspace{-5pt}\tikzmk{B}\boxit{green}

\end{algorithmic}
\end{algorithm}

In Section~\ref{sec: results} we demonstrate a significant decrease in the uncertainty volume, as defined in Equation \eqref{eq: uncertainty volume} for each procedure, whether applied globally or locally, compared to prior work. 
On the one hand, the E-PUQ procedure is the simplest and can be applied locally to any task, 
and globally to certain tasks where the computation of $d$ PCs is feasible.
On the other hand, the DA-PUQ and RDA-PUQ procedures are more involved and can be applied both globally or locally to any task, while these are particularly effective in cases in which pixels exhibit strong correlations, such as in the image colorization task.
Our method is visually illustrated in Figure~\ref{fig: calibration}, showing a sampling methodology and a calibration scheme using the full PCs or only a subset of them.

\subsection{Diffusion Models for the Approximation Phase}
\label{sec: our implementation}

\noindent The approximation phase, summarized in Algorithm~\ref{alg: general scheme} in \textcolor{pink}{RED}, can be achieved in various ways. In this section, we describe the implementation we used to obtain the results in Section~\ref{sec: results}. While we aim to construct the uncertainty axes and intervals in the most straightforward way, further exploration of more advanced methods to achieve the PCs is left for future work.

In our implementation, we leverage the recent advances in stochastic regression solvers for inverse problems based on 
diffusion models, which enable to train a machine learning model to generate high-quality samples 
from $\thP$.
Formally, we define $f_\theta : \mathcal{X} \times \mathcal{Z} \rightarrow \mathcal{Y}$ as a stochastic regression solver for an inverse problem in global mode, where $\mathcal{Z}$ is the noise seed space. Similarly, in local mode, we consider $f_\theta : \mathcal{X} \times \mathcal{Z} \rightarrow \mathcal{Y}_{\text{patch}}$.
Given an input instance $x \in \dR^d$, we propose to generate $K$ samples, denoted by $\{f_\theta(x,z_i)\}_{i=1}^K$, where, $f_\theta(x,z_i) \sim \thP$.
These samples are used to estimate the PCs of possible solutions and their importance weights using the SVD decomposition of the generated samples.
The importance weights assign high values to axes with large variance among projected samples, and low ones to those with small variance.
In Section~\ref{sec: calibration}, we elaborate on how these weights are used in the calibration phase.
Additionally, the samples are utilized to estimate the conditional mean, $\hat{\mu}(x)$, and the lower and upper bounds, $\tilde{l}(x)$ and $\tilde{u}(x)$, necessary for Equation~\eqref{eq: intervals}.
$\tilde{l}(x)$ and $\tilde{u}(x)$ are obtained by calculating quantiles of the projected samples onto each PC, with a user-specified miss-coverage ratio $\alpha \in (0,1)$.

To capture the full spread and variability of $\thP$, it is necessary to generate at least $K=d$ samples to feed to the SVD procedure, which is computationally challenging for high-dimensional data.
As a way out, we suggest working locally on patches, where $d$ is small and fully controlled by the user by specifying the patch size to work on.
However, for tasks with strong pixel correlation, such as image colorization, a few PCs can describe the variability of $\thP$ with a very small error. Therefore, only a few samples (i.e., $K \ll d$) are required for the SVD procedure to construct meaningful PCs for the entire image, while capturing most of the richness in $\thP$.
We formally summarize our sampling-based methodology, in either global or local modes, in Algorithm~\ref{alg: approximation phase}.

\begin{algorithm}[htb]
\small
\caption{Approximation Phase 
via Sampling}
\label{alg: approximation phase}
\begin{algorithmic}[1]

\Require{Instance $x \in \mathcal{X}$. Conditional stochastic generative model $f_\theta : \mathcal{X} \rightarrow \mathcal{Y}$ or $f_\theta : \mathcal{X} \rightarrow \mathcal{Y}_{\text{patch}}$. 
Maximal PCs / samples number $K \leq d$. Misscoverage ratio $\alpha \in (0,1)$.}


\Comment{Generate samples drawn from $\thP$}

\For{$i = 1$ to $K$}

\State $\hat{y}_i(x) \gets f_\theta(x,z_i)$

\EndFor

\Comment{Compute conditional mean}

\State $\hat{\mu}(x) \gets \frac{1}{K} \sum_{i=1}^K \hat{y}_i(x)$

\Comment{Apply SVD decomposition and extract the PCs and weights}

\State $\hat{Y}(x) \gets [ \hat{y}_1(x) , \hat{y}_2(x) \dots \hat{y}_K(x) ] \in \dR^{d \times K}$

\State $\hat{Y}(x) - \hat{\mu}(x) \cdot \mathbf{1}_K^T = \hat{V}(x) \hat{\Sigma}(x) \hat{U}(x)^T$

\State $\hat{B}(x) \gets \{ \hat{v}_1(x) , \hat{v}_2(x) \dots \hat{v}_K(x) \}$, where $\hat{v}_i(x) = [\hat{V}(x)]_i$

\State $\hat{w}(x) \gets \left[ \hat{\sigma}_1(x)^2, \dots, \hat{\sigma}_K(x)^2 \right] /c \in \dR^K $, where $\hat{\sigma}_i(x) = [\hat{\Sigma}(x)]_i$ and $c=\sum_{j=1}^{K}{\hat{\sigma}_j(x)}^2$.

\Comment{Compute $\alpha/2$ and $1-\alpha/2$ empirical quantiles of projected samples onto each PC}

\For{$i = 1$ to $K$}

\State $\tilde{l}(x)_i \gets \hat{\mathcal{Q}}_{\alpha/2} ( \{ \hat{v}_i(x)^T (\hat{y}_j(x)-\hat{\mu}(x)) \}_{j=1}^K ) $

\State $\tilde{u}(x)_i \gets \hat{\mathcal{Q}}_{1 - \alpha/2} ( \{ \hat{v}_i(x)^T (\hat{y}_j(x) -\hat{\mu}(x)) \}_{j=1}^K ) $

\EndFor

\Ensure{$K$ PCs $\hat{B}(x)$, importance weights $\hat{w}(x)$, conditional mean $\hat{\mu}(x)$, lower and upper bounds $\tilde{l}(x)$ and $\tilde{u}(x)$.}

\end{algorithmic}
\end{algorithm}

\subsection{Calibration Phase} 
\label{sec: calibration}

\noindent In order to refine the approximation phase and obtain valid uncertainty axes and intervals that satisfy the guarantees of Equation~\eqref{eq: theoretical coverage gaurentee} and Equation~\eqref{eq: theoretical reconstruction gaurentee}, it is necessary to apply a calibration phase, as summarized in Algorithm~\ref{alg: general scheme} in \textcolor{blue2}{BLUE}. This phase includes two different options based on particular conditions on the number of PCs to be constructed and maintained during the calibration procedure or during inference, when applied either globally or locally. Below we outline each of these options in more details.

\subsubsection{Exact PUQ}
\label{sec: Exact PUQ}

The \emph{Exact PUQ} (E-PUQ) procedure 
provides the complete uncertainty of the $d$-dimensional posterior distribution, $\tP$. 
In this case, the exact reconstruction property discussed in Section~\ref{sec: Problem Formulation} is satisfied, and Equation~\eqref{eq: theoretical reconstruction gaurentee} is fulfilled with $0\%$ error ($\beta=0$) across $100\%$ ($q=1.0$) of the pixels. Therefore, the calibration is simple, involving only a scaling of intervals to ensure Equation~\eqref{eq: theoretical coverage gaurentee} is satisfied with high probability, similar to previous work~\cite{angelopoulos2022image,horwitz2022conffusion}.

Formally, for each input instance $x$ and its corresponding ground-truth value $y \in \mathbb{R}^d$ in the calibration data, we use the estimators obtained in the approximation phase to get $d$ PCs of possible solutions $\hat{B}(x)$, their corresponding importance weights $\hat{w}(x)$, the conditional mean $\hat{\mu}(x)$, and the lower and upper bounds, denoted by $\tilde{l}(x)$ and $\tilde{u}(x)$. We then define the scaled intervals to be those specified in Equation~\eqref{eq: intervals}, with the upper and lower bounds defined as $\hat{u}(x) := \lambda \tilde{u}(x)$ and $\hat{l}(x) := \lambda \tilde{l}(x)$, where $\lambda \in \dR^+$ is a tunable parameter that controls the scaling. Notably, the size of the uncertainty intervals decreases as $\lambda$ decreases. We denote the scaled uncertainty intervals by $\mathcal{T}_\lambda(x; \hat{B}(x))$. 
The following weighted coverage loss function is used to guide our design of $\lambda$: 
{\small\begin{equation}
\label{eq: full coverage loss}
\mathcal{L} (x, y; \lambda) := \sum_{i=1}^{d} \hat{w}_i(x) \cdot \mathbf{1}\left\{\hat{v}_i(x)^T y \not\in \mathcal{T}_\lambda(x; \hat{B}(x))_i \right\}. 
\end{equation}}
This loss is closely related to the expression in Equation~\eqref{eq: theoretical coverage gaurentee}, and while it may seem arbitrary at first, this choice is a direct extension to the one practiced in ~\cite{angelopoulos2022image,horwitz2022conffusion}. In Appendix \ref{app: coverage loss justification} we provide an additional justification for it, more tuned to the realm discussed in this paper.

Our goal is to ensure that the  expectation of $\mathcal{L} (x, y; \lambda)$ is below a pre-specified threshold, $\alpha$, with high probability over the calibration data. 
This is accomplished by a conformal prediction based calibration scheme, and in our paper we use the LTT \cite{angelopoulos2021learn} procedure, which guarantees the following:
{\small\begin{equation}
\label{eq: Exact PUQ guarantee}
\dP \left( \dE[\mathcal{L} (x, y; \hat{\lambda})] \leq \alpha \right) \geq 1 - \delta ~,
\end{equation}}
for a set of candidate values of $\lambda$, given as the set $\hat{\Lambda}$. 
$\delta \in (0,1)$ is an error level on the calibration set and $\hat{\lambda}$ is the smallest value within $\hat{\Lambda}$ 
satisfying the above condition, so as to provide the smallest uncertainty volume over the scaled intervals, as defined in Equation~\eqref{eq: uncertainty volume}, which we denote by $\mathcal{V}_{\hat{\lambda}}$. 

Put simply, the above guarantees that more than $1 - \alpha$ of the ground-truth values projected onto the full $d$ PCs of $\thP$ are contained in the uncertainty intervals with probability at least $1 - \delta$, where the latter probability is over the randomness of the calibration set. The scaling factor takes into account the weights to ensure that uncertainty intervals with high variability contain a higher proportion of projected ground-truth values than those with low variability. This is particularly important for tasks with strong pixel correlations, where the first few PCs capture most of the variability in possible solutions.
We describe in detail the E-PUQ procedure in Algorithm~\ref{alg: exact puq}.

\begin{algorithm}[htb]
\small
\caption{Exact PUQ Procedure}
\label{alg: exact puq}
\begin{algorithmic}[1]

\Require{Calibration set $\mathcal{S}_\text{cal} := \{ x_i , y_i \}_{i=1}^n$. Scanned calibration parameter values $\Lambda = [ 1 \dots \lambda_{\text{max}} ]$.
Approximation phase estimations $\hat{B},\hat{w},\hat{\mu},\tilde{u},\tilde{l}$. Misscoverage ratio $\alpha \in (0,1)$. Calibration error level $\delta \in (0,1)$.}



\For{$(x,y) \in \mathcal{S}_\text{cal}$}

\State $\hat{B}(x),\hat{w}(x),\hat{\mu}(x),\tilde{u}(x),\tilde{l}(x) \gets$
Apply Algorithm~\ref{alg: approximation phase} to \par\hspace*{3em} $x$, with the choice of $K = d$ samples

\For{$\lambda \in \Lambda$}

\Comment{Scale uncertainty intervals}

\State $\hat{u}(x) \gets \lambda \tilde{u}(x)$ and $\hat{l}(x) \gets \lambda \tilde{l}(x)$

\State $\mathcal{T}_\lambda(x; \hat{B}(x)) \gets$ Equation~\eqref{eq: intervals} using $\hat{\mu}(x),\hat{u}(x),\hat{l}(x)$

\Comment{Compute weighted coverage loss, Equation~\eqref{eq: full coverage loss}}

\State $\mathcal{L}(x,y;\lambda) \gets$\\
\hspace*{4em}\smash{$\sum_{i=1}^{d} \hat{w}_i(x) \cdot \mathbf{1}\left\{{\hat{v}_i(x)}^T y \not\in \mathcal{T}_\lambda(x; \hat{B}(x))_i \right\}$}

\EndFor

\EndFor

\State $\hat{\Lambda} \gets$ Extract valid $\lambda$s from LTT \cite{angelopoulos2021learn} applied on $ \{ \mathcal{L}(x,y;\lambda) : {(x,y) \in S_{\text{cal}}, \lambda \in \Lambda}\} $ at risk level $\alpha$ and error level $\delta$, referring to Equation~\eqref{eq: Exact PUQ guarantee}.

\Comment{Compute the minimizer for the uncert. volume, Equation~\eqref{eq: uncertainty volume}}

\State $\hat{\lambda} \gets \arg \min_{\lambda \in \hat{\Lambda}} \left\{ \frac{1}{n} \sum_{i=1}^n \mathcal{V}_\lambda(x_i;\hat{B}(x)) \right\}$

\Ensure{Given a new instance $x \in \mathcal{X}$, 
obtain valid uncertainty intervals for it,  $\mathcal{T}_{\hat{\lambda}}(x; \hat{B}(x))$.}


\end{algorithmic}
\end{algorithm}

\subsubsection{Dimension-Adaptive PUQ}
\label{sec: Dimensional Adaptive PUQ}

The E-PUQ procedure assumes the ability to construct and maintain $d$ PCs, which can be computationally challenging both locally and globally.
Furthermore, an uncertainty quantification over these axes may be less intuitive, due to the many axes involved, thus harming the method's interpretability (see discussion in Section~\ref{sec: Problem Formulation}).
To address these, we propose the \emph{Dimension-Adaptive PUQ} (DA-PUQ) procedure, which describes the uncertainty region with fewer axes, $K \leq d$. 
The use of only a few leading dimensions, e.g., $K=3$, can lead to a more interpretable uncertainty region, enabling an effective visual navigation within the obtained uncertainty range.

While this approach does not satisfy the exact reconstruction property (see Section~\ref{sec: Problem Formulation}), the decomposed ground-truth values can still be recovered through the $K$ PCs with a small user-defined error in addition to the coverage guarantee. By doing so, we can achieve both the guarantees outlined in Equation~\eqref{eq: theoretical coverage gaurentee} and Equation~\eqref{eq: theoretical reconstruction gaurentee} with high probability.




To satisfy both the coverage and reconstruction guarantees while enhancing interpretability, we use a dynamic function, $\hat{k}(x): \mathcal{X} \rightarrow \dN$, and a scaling factor to control the reconstruction and coverage risks.
The function $\hat{k}(x)$ determines the number of top PCs (out of $K$) to include in the uncertainty region, focusing on  the smallest number that can satisfy both Equation~\eqref{eq: theoretical coverage gaurentee} and \eqref{eq: theoretical reconstruction gaurentee}, so as to increase interpretability.

Formally, for each input instance $x$ and its corresponding ground-truth value $y \in \mathbb{R}^d$ in the calibration data, we use the estimators obtained in the approximation phase to estimate $K \leq d$ PCs of possible solutions, denoted by $\hat{B}(x)$, their corresponding importance weights, denoted by $\hat{w}(x)$, the conditional mean denoted by $\hat{\mu}(x)$, and the lower and upper bounds denoted by $\tilde{l}(x)$ and $\tilde{u}(x)$, respectively.
We then introduce a threshold $\lambda_1 \in (0,1)$ for the decay of the importance weights over the PCs of solutions to $x$.
The adaptive number of PCs to be used is defined as follows:
{\small\begin{equation}
\label{eq: adaptive k}
\hat{k}(x; \lambda_1) := \min_{1\le k\le K} \left\{ k ~~~\mbox{s.t.}~~~ \sum_{i=1}^k \hat{w}_i(x) \geq \lambda_1 \right\} .
\end{equation}}
Obviously, the importance weights are arranged in a descending order, starting from the most significant axis and ending with the least significant one. 
Furthermore, let $q \in (0,1)$ be a specified ratio of pixels, and $\beta \in (0,1)$ be a maximum allowable reconstruction error over this ratio. The reconstruction loss function to be controlled is defined as:
{\small\begin{align}
\label{eq: interpertable reconstruction loss}
\mathcal{L}_1(x,y;\lambda_1) := \hat{\mathcal{Q}}_q \left( \left\{ \left| \sum_{j=1}^{\hat{k}(x;\lambda_1)} \hat{v}_j(x)^T y_c \hat{v}_j(x) - y_c \right|_i \right\}_{i=1}^d \right) ~,
\end{align}} 
where $\hat{\mathcal{Q}}_q(\cdot)$ selects the empirical $q$-quantile of the reconstruction errors, and $y_c= y-{\hat \mu}(x)$ is the ground-truth image centered around ${\hat \mu}(x)$.
In Appendix~\ref{app: reconstruction loss justification}, we discuss further this specific loss function for controlling the capability of the linear subspace to capture the richness of the complete $d$-dimensional posterior distribution.

At the same time, we also control the coverage risk over the $\hat{k}(x)$ PCs, with $\alpha \in (0,1)$ representing a user-specified acceptable misscoverage rate and $\lambda_2 \in \dR^+$ representing the calibration factor parameter. To control this coverage risk, we define the coverage loss function to be the same as in Equation~\eqref{eq: full coverage loss}, but limited to the $\hat{k}(x)$ PCs, that is:
{\small\begin{align}
\label{eq: dimensional adaptive loss}
\mathcal{L}_2 (x&, y; \lambda_1, \lambda_2) := \\
\nonumber &\sum_{i=1}^{\hat{k}(x;\lambda_1)} \hat{w}_i(x) \cdot \mathbf{1}\left\{\hat{v}_i(x)^T y \not\in \mathcal{T}_{\lambda_2}(x; \hat{B}(x))_i \right\} .
\end{align}}

Finally, using the reconstruction loss function of Equation~\eqref{eq: interpertable reconstruction loss} and the coverage loss function of Equation~\eqref{eq: dimensional adaptive loss}, we seek to minimize the uncertainty volume, defined in Equation~\eqref{eq: uncertainty volume}, for the scaled intervals where any unused axes (out of $d$) are fixed to zero. We denote this uncertainty volume as $\mathcal{V}_{\lambda_1,\lambda_2}$.
The minimization of $\mathcal{V}_{\lambda_1,\lambda_2}$ is achieved by minimizing $\lambda_1$ and $\lambda_2$, while ensuring that the guarantees of Equation~\eqref{eq: theoretical coverage gaurentee} and Equation~\eqref{eq: theoretical reconstruction gaurentee} hold with high probability over the calibration data.
This can be provided, for example, through the LTT \cite{angelopoulos2021learn} calibration scheme, which guarantees the following:
{\small\begin{align}
\label{eq: Dimensional Adaptive PUQ guarantee}
\dP \left( 
\begin{array}{c}
\dE[\mathcal{L}_1 (x, y; \hat{\lambda}_1)] \leq \beta \\ \dE[\mathcal{L}_2 (x, y; \hat{\lambda}_1, \hat{\lambda}_2)] \leq \alpha 
\end{array}
\right) \geq 1 - \delta ~,
\end{align}}
where $\hat{\lambda}_1$ and $\hat{\lambda}_2$ are the minimizers for the uncertainty volume among valid calibration parameter results, $\hat{\Lambda}$, obtained through the LTT procedure.
In other words, we can reconstruct a fraction $q$ of the ground-truth pixel values with an error no greater than $\beta$, and a fraction of more than $1-\alpha$ of the projected ground-truth values onto the first $\hat{k}(x;\hat{\lambda}_1)$ PCs of $\tP$ are contained in the uncertainty intervals, with a probability of at least $1-\delta$.
A detailed description of the DA-PUQ procedure is given in Algorithm~\ref{alg: dimensional adaptive puq}.

The above-described DA-PUQ procedure reduces the number of PCs to be constructed to $K \leq d$ while using $\hat{k}(x;\hat{\lambda}_1) \leq K$ PCs, leading to increased efficiency in both time and space during inference.
However, determining manually the smallest $K$ value that can guarantee both Equation~\eqref{eq: theoretical coverage gaurentee} and Equation~\eqref{eq: theoretical reconstruction gaurentee} can be challenging.
To address this, we propose an expansion of the DA-PUQ procedure; the \emph{Reduced Dimension-Adaptive PUQ} (RDA-PUQ) procedure that also controls the maximum number of PCs, $K$, required for the uncertainty assessment.
This approach is advantageous for inference as it reduces the number of samples required to construct the PCs using Algorithm~\ref{alg: approximation phase}, while ensuring both the coverage and reconstruction guarantees of Equation~\eqref{eq: theoretical coverage gaurentee} and Equation~\eqref{eq: theoretical reconstruction gaurentee} with high probability.
The RDA-PUQ procedure is fully described in Appendix~\ref{app: Reduced Dimensional Adaptive PUQ}.

\begin{algorithm}[htb]
\small
\caption{Dimension-Adaptive PUQ Procedure}
\label{alg: dimensional adaptive puq}
\begin{algorithmic}[1]

\Require{
Calibration set $\mathcal{S}_{\text{cal}} := \{ x_i , y_i \}_{i=1}^n$. Scanned calibration parameter values $\Lambda^1 \gets [ 1 \dots {\lambda_1}_{\text{max}} ]$ and $\Lambda^2 \gets [ 1 \dots {\lambda_2}_{\text{max}} ]$.
Maximal PCs number $K \leq d$. Approximation phase estimators $\hat{B},\hat{w},\hat{\mu},\tilde{u},\tilde{l}$. Recovered pixels ratio $q \in (0,1)$. Reconstruction error $\beta \in (0,1)$. Misscoverage ratio $\alpha \in (0,1)$. Calibration error level $\delta \in (0,1)$. 
For an effective calibration, $\alpha, \beta, \delta$ should be close to $0$ while $q$ should be close to $1$.
}


\For{$(x,y) \in \mathcal{S}_\text{cal}$}

\State $\hat{B}(x),\hat{w}(x),\hat{\mu}(x),\tilde{u}(x),\tilde{l}(x) \gets$ Apply Algorithm~\ref{alg: approximation phase} to \par\hspace*{3em} $x$, with the choice of $K$ samples

\For{$\lambda_1 \in \Lambda_1$}

\Comment{Compute adaptive dimensionality, Equation~\eqref{eq: adaptive k}}

\State $\hat{k}(x;\lambda_1) \gets \min_k \left\{ k : \sum_{i=1}^K \hat{w}_i(x) \geq \lambda_1 \right\} $

\Comment{Compute reconstruction loss, Equation~\eqref{eq: interpertable reconstruction loss}}

\State $y_c \gets y - \hat{\mu}(x)$

\State $\mathcal{L}_1(x,y;\lambda_1) \gets $\par
\hspace*{-1em}$\hat{\mathcal{Q}}_q \bigg( \bigg\{ \bigg| \sum_{j=1}^{\hat{k}(x;\lambda_1)} \hat{v}_j(x)^T y_c \hat{v}_j(x) - y_c \bigg|_i \bigg\}_{i=1}^d \bigg)$

\For{$\lambda_2 \in \Lambda_2$}

\Comment{Scale uncertainty intervals}

\State $\hat{u}(x_i) \gets \lambda_2 \tilde{u}(x)$ and $\hat{l}(x) \gets \lambda_2 \tilde{l}(x)$

\State $\mathcal{T}_{\lambda_2}(x; \hat{B}(x)) \gets$ Eq.~\eqref{eq: intervals} using $\hat{\mu}(x),\hat{u}(x),\hat{l}(x)$

\Comment{Compute weighted coverage loss, Equation~\eqref{eq: full coverage loss}}

\State $\mathcal{L}_2(x,y;\lambda_1,\lambda_2) \gets \sum_{i=1}^{\hat{k}(x;\lambda_1)} \hat{w}_i(x) \cdot $\par
\hspace*{4em}\smash{$ \mathbf{1}\left\{{\hat{v}_i(x)}^T y \not\in \mathcal{T}_{\lambda_2}(x; \hat{B}(x))_i \right\}$}

\EndFor

\EndFor

\EndFor

\State $\hat{\Lambda} \gets$ Extract valid $\lambda$s from LTT \cite{angelopoulos2021learn} applied on $\{ ( \mathcal{L}_1(x,y;\lambda_1) , \mathcal{L}_2(x,y;\lambda_1,\lambda_2) ) : {(x,y) \in S_{\text{cal}}, \lambda_1 \in \Lambda^1 , \lambda_2 \in \Lambda^2}\}$ at risk levels ($\beta,\alpha$) and error level $\delta$, referring to Equation~\eqref{eq: Dimensional Adaptive PUQ guarantee}

\Comment{Compute the minimizers for the uncer. volume, Equation~\eqref{eq: uncertainty volume}}

\State $\hat{\lambda}_1, \hat{\lambda}_2 \gets \arg \min_{\lambda_1,\lambda_2 \in \hat{\Lambda}} \left\{ \frac{1}{n} \sum_{i=1}^n \mathcal{V}_{\lambda_1,\lambda_2}(x_i;\hat{B}(x_i)) \right\}$

\Ensure{Given a new instance $x \in \mathcal{X}$, obtain valid uncertainty intervals for it,  $\mathcal{T}_{\hat{\lambda}_2}(x; \hat{B}(x))$ over $\hat{k}(x;\hat{\lambda}_1) \leq K$ PCs.}

\end{algorithmic}
\end{algorithm}


\section{Empirical Study}
\label{sec: results}

\begin{figure*}[t]
    \centering
    {\hspace{-6cm}\small $x$ ~~~~~~~~~~ $y$ ~~~~~~~~~~~~~~~~~~~~~~~~~~~~~~~~~~~~~~~~~~~~~~~~~~~~~~~~~~~~~~~~~~~ Samples }
    \includegraphics[width=1.0\textwidth,trim={1cm 1cm 1cm 1cm},clip]{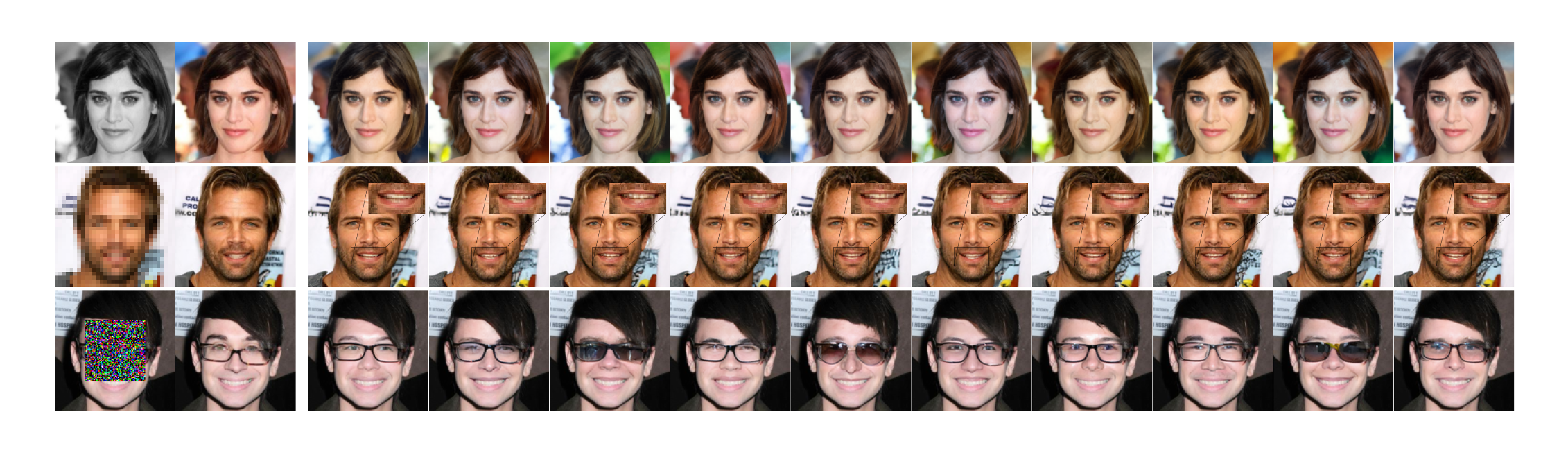}
    \caption{The three image recovery tasks, colorization (top), super-resolution (middle) and inpainting (bottom). For each we present a given measurement $x$, the ground-truth $y$, and $10$ candidate samples from the (approximated) posterior distribution. These samples fuel the approximation phase in our work.}
    \label{fig: samples}
\end{figure*}

\noindent This section presents a comprehensive empirical study of our proposed method PUQ, applied to three challenging tasks: image colorization, super-resolution, and inpainting, over the CelebA-HQ dataset \cite{karras2017progressive}. Our approximation phase starts with a sampling from the posterior, applied in our work by the SR3 conditional diffusion model~\cite{saharia2022image}. Figure~\ref{fig: samples} presents typical sampling results for these three tasks, showing the expected diversity in the images obtained. 

The experiments we present herein verify that our method satisfies both the reconstruction and coverage guarantees and demonstrate that PUQ provides more confined uncertainty regions compared to prior work, including im2im-uq \cite{angelopoulos2022image} and Conffusion \cite{horwitz2022conffusion}.
Through the experiments, we present superiority in uncertainty volume, as defined in Equation~\eqref{eq: uncertainty volume}, and in interpretability through the use of only a few PCs to assess the uncertainty of either a patch or a complete image.
All the experiments were conducted over 100 calibration-test splits. For in-depth additional details of our experiments and the settings used, we refer the reader to Appendix~\ref{app: Experimental Details}. Additionally, an ablation study has been conducted, as elaborated in Appendix~\ref{app: ablation study}. This study presents an analysis of user-defined parameters: $\alpha$, $\beta$, $q$, and $\delta$, aiming to provide a comprehensive insight into their selection.
Furthermore, we have investigated the trade-off between precision and complexity in Appendix~\ref{app: tradeoff} to offer a complete understanding of our method's performance.

\subsection{Evaluation Metrics}
\label{sec: evaluation metrics}


\noindent Before presenting the results, we discuss the metrics used to evaluate the performance of the different methods. Although our approach is proved to guarantee Equation~\eqref{eq: Exact PUQ guarantee} for E-PUQ and Equation~\eqref{eq: Dimensional Adaptive PUQ guarantee} for DA-PUQ, 
(through LTT \cite{angelopoulos2021learn}), we assess the validity and tightness of these guarantees as well.

\textbf{Empirical coverage risk.} ~ We measure the risk associated with the inclusion of projected unseen ground-truth values in the uncertainty intervals. In E-PUQ, we report the average coverage loss, defined in Equation~\eqref{eq: full coverage loss}. In the case of DA-PUQ and RDA-PUQ, we report the value defined by Equation~\eqref{eq: dimensional adaptive loss}.

\textbf{Empirical reconstruction risk.} ~ We measure the risk in recovering unseen ground-truth pixel values using the selected PCs. In the case of E-PUQ, this risk is zero by definition. However, for DA-PUQ and RDA-PUQ, we report the average reconstruction loss, defined by Equation~\eqref{eq: interpertable reconstruction loss}.

\textbf{Interval-Size.} ~ We report the calibrated uncertainty intervals' sizes of Equation~\eqref{eq: intervals}, and compare them with baseline methods. For E-PUQ, we compare intervals over the full basis set of PCs with the intervals in the pixel domain used in previous work.
In the DA-PUQ and RDA-PUQ procedures, we apply dimensionality reduction to $K \ll d$ dimensions. To validly compare the intervals' sizes of these methods to those methods over the full $d$ dimensions, we pad the remaining $d-K$ dimensions with zeros as we assume that the error in reconstructing the ground-truth from the dimensionally reduced samples is negligible.


\textbf{Uncertainty Volume.} We report these volumes, defined in Equation~\eqref{eq: uncertainty volume}, for the calibrated uncertainty regions and compare them with previous work. A smaller  volume implies a higher level of certainty in probable solutions to $\tP$. In E-PUQ, we compare volumes over the full basis set of PCs, whereas for the DA-PUQ and RDA-PUQ procedures, we pad the remaining dimensions with zeros. 


\subsection{Local Experiments on Patches}

\begin{table}[]
\centering
\caption{\textbf{Local Experiments:} Quantitative comparison of the means and standard deviations of our locally applied PUQ method on RGB patch resolution of 8x8, utilizing the two proposed procedures. Note that in this experiment $d=8\times 8\times 3 = 192$ and $\epsilon=1e-10$ for the volume computation.}
\label{tab: local results}
\resizebox{\columnwidth}{!}{%
\begin{tabular}{@{}llccc@{}}
\toprule
                                       &                     & \textbf{Recons. Risk} & \textbf{Dim. $\hat{k}(x)$ / $K$} & \textbf{Uncert. Volume} \\ \midrule
\multirow{5}{*}{\textbf{Colorization}} & im2im-uq \cite{angelopoulos2022image}   & $0$                  & $192$ / $192$          & $\expnumber{1.6}{-1} \pm \expnumber{7.2}{-2}$                   \\
                                       & Conffusion \cite{horwitz2022conffusion} & $0$                  & $192$ / $192$          & $\expnumber{1.7}{-1} \pm \expnumber{1.3}{-1}$                   \\
                                       & \textbf{E-PUQ}  & $0$                  & $192$ / $192$          & $\expnumber{2.3}{-3} \pm \expnumber{9.8}{-4}$                   \\
                                       & \textbf{DA-PUQ}      & $\expnumber{2.5}{-2} \pm \expnumber{5.3}{-4}$                  & $1.6 \pm 0.77$ / $100$          & $\expnumber{2.4}{-11} \pm \expnumber{1.3}{-11}$                   \\
                                       & \textbf{RDA-PUQ}     & $\expnumber{1.7}{-2} \pm \expnumber{9.7}{-4}$                  & $3.8 \pm 1.8$ / $11.8 \pm 6.4$         & $\expnumber{6.8}{-11} \pm \expnumber{3.9}{-11}$                   \\ \midrule
\multirow{5}{*}{\textbf{\begin{tabular}[c]{@{}l@{}}Super-\\ Resolution\end{tabular}}} & im2im-uq \cite{angelopoulos2022image} & $0$ & $192$ / $192$ & $\expnumber{8.8}{-2} \pm \expnumber{5.7}{-2}$ \\
                                       & Conffusion \cite{horwitz2022conffusion} & $0$                  & $192$ / $192$          & $\expnumber{8.8}{-2} \pm \expnumber{6.9}{-2}$                   \\
                                       & \textbf{E-PUQ}  & $0$                  & $192$ / $192$          & $\expnumber{1.3}{-2} \pm \expnumber{7.5}{-3}$                   \\
                                       & \textbf{DA-PUQ}      & $\expnumber{2.5}{-2} \pm \expnumber{3.4}{-4}$                  & $11.1 \pm 5.5$ / $192$          & $\expnumber{3.4}{-10} \pm \expnumber{3.5}{-10}$                   \\
                                       & \textbf{RDA-PUQ}     & $\expnumber{2.0}{-2} \pm \expnumber{0.0}{-0}$                  & $22.8 \pm 6.8$ / $70.0 \pm 0.0$          & $\expnumber{1.6}{-9} \pm \expnumber{1.3}{-9}$                   \\ \midrule
\multirow{5}{*}{\textbf{Inpainting}}   & im2im-uq \cite{angelopoulos2022image}   & $0$                  & $192$ / $192$          & $\expnumber{2.8}{-1} \pm \expnumber{1.4}{-1}$                   \\
                                       & Conffusion \cite{horwitz2022conffusion} & $0$                  & $192$ / $192$          & $\expnumber{2.7}{-1} \pm \expnumber{1.6}{-1}$                   \\
                                       & \textbf{E-PUQ}  & $0$                  & $192$ / $192$          & $\expnumber{1.9}{-2} \pm \expnumber{1.0}{-2}$                   \\
                                       & \textbf{DA-PUQ}      & $\expnumber{1.8}{-2} \pm \expnumber{1.0}{-4}$                  & $39.4 \pm 11.8$ / $192$          & $\expnumber{3.6}{-8} \pm \expnumber{7.3}{-8}$                   \\
                                       & \textbf{RDA-PUQ}     & $\expnumber{1.8}{-2} \pm \expnumber{1.6}{-3}$                  & $55.6 \pm 10.2$ / $72.6 \pm 30.3$          & $\expnumber{1.3}{-7} \pm \expnumber{8.8}{-8}$                   \\ \bottomrule
\end{tabular}%
}
\end{table}

\begin{figure}[t]
    \centering
    \includegraphics[width=\columnwidth,trim={0 0.1cm 0.1cm 0},clip]{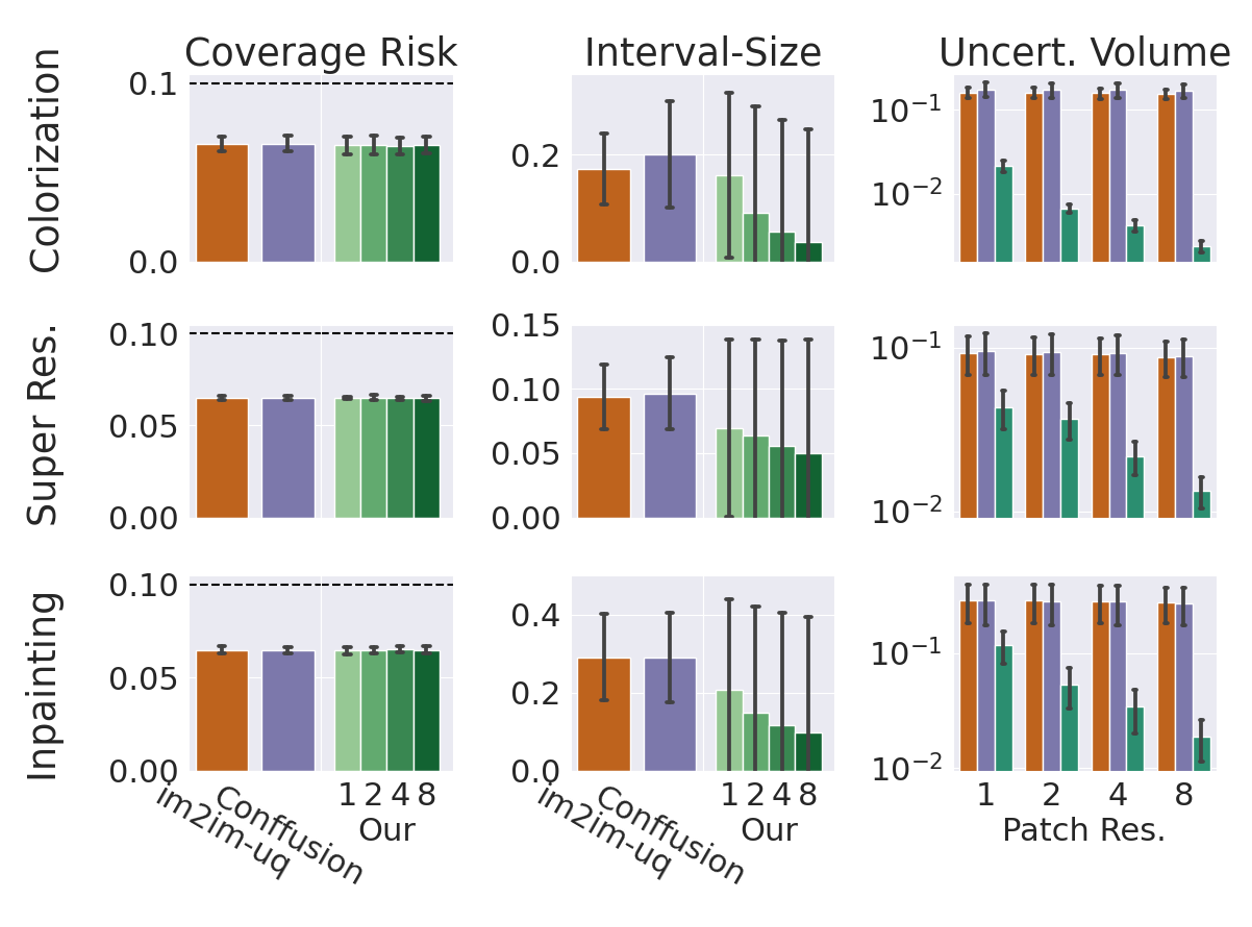}
    \caption{\textbf{Local Experiments:} A comparison of E-PUQ (see Section~\ref{sec: Exact PUQ}) with previous work -- im2im-uq \cite{angelopoulos2022image} and Conffusion \cite{horwitz2022conffusion}. These methods are applied locally on patches with $\alpha = \delta = 0.1$. Each column corresponds to a relevant metric (see Section~\ref{sec: evaluation metrics}), and each row corresponds to a specific task. The uncertainty volume was computed with $\epsilon = 1e-10$. Results indicate that our approach achieves superior uncertainty volume.}
    \label{fig: local exact puq results}
\end{figure}

\begin{figure}[t]
    \centering
    \includegraphics[width=1.1\columnwidth]{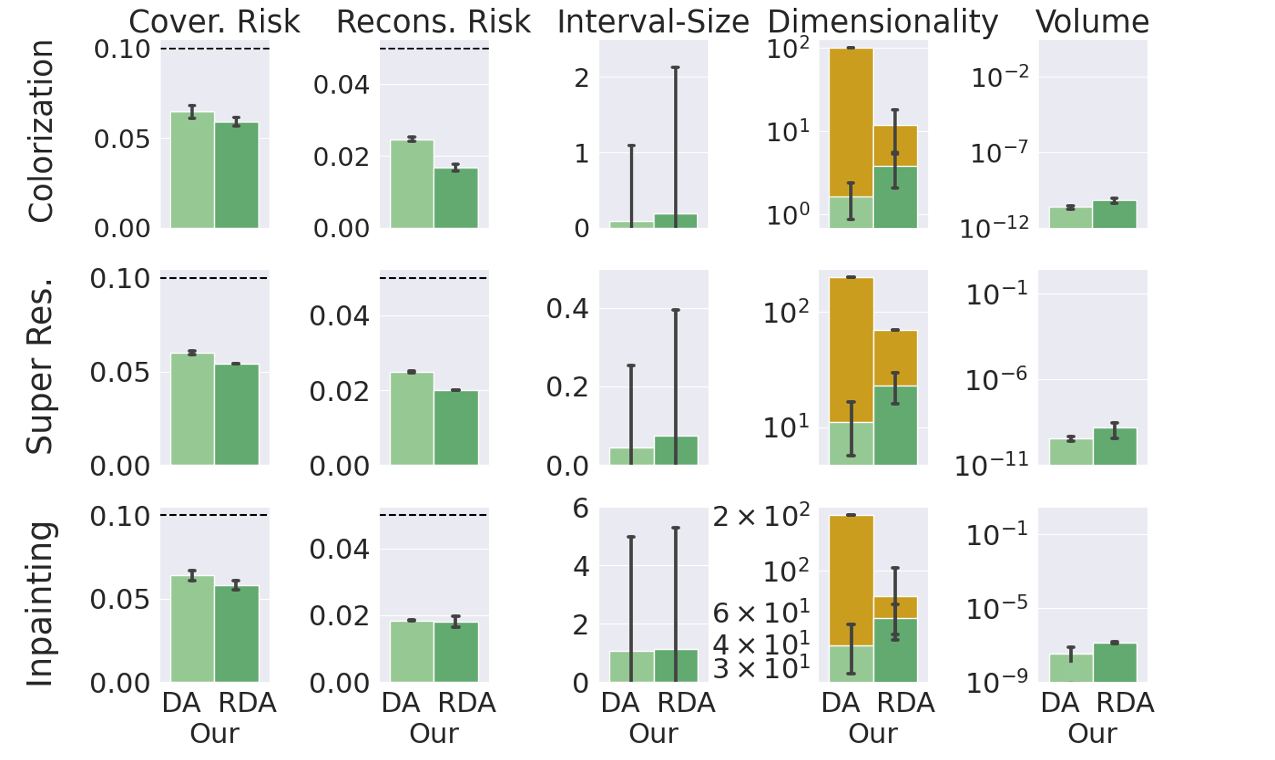}
    \caption{\textbf{Local Experiments:} A comparison of DA-PUQ (see Section~\ref{sec: Dimensional Adaptive PUQ}) and RDA-PUQ (see Appendix~\ref{app: Reduced Dimensional Adaptive PUQ}), when applied locally on 8x8 patches with  $\alpha=\delta=0.1$, $\beta = 0.05$ and $q=0.9$.
    Each column corresponds to a relevant metric (see Section~\ref{sec: evaluation metrics}), and each row corresponds to a specific task.
    The uncertainty volume was computed with $\epsilon = 1e-10$.
    Here, the dimensionality is presented by two overlapping bars, where the yellow bars represent the distribution of $K$ in DA-PUQ and $\hat{K}$ in RDA-PUQ, and the inner bars represent the distribution of $\hat{k}(x)$ in both cases.}
    \label{fig: local adaptive and reduced puq results}
\end{figure}

\begin{figure*}[t]
    \centering
    {\hspace*{-1.3cm}\small $x$ ~~~~~~~~~~~ $y$ ~~~~~~~~ im2im-uq ~~~ Conffusion ~~~ Our-1x1x3~ Our-2x2x3~ Our-4x4x3~ Our-8x8x3}
    \includegraphics[width=0.9\textwidth,trim={1cm 1cm 1cm 1cm},clip]{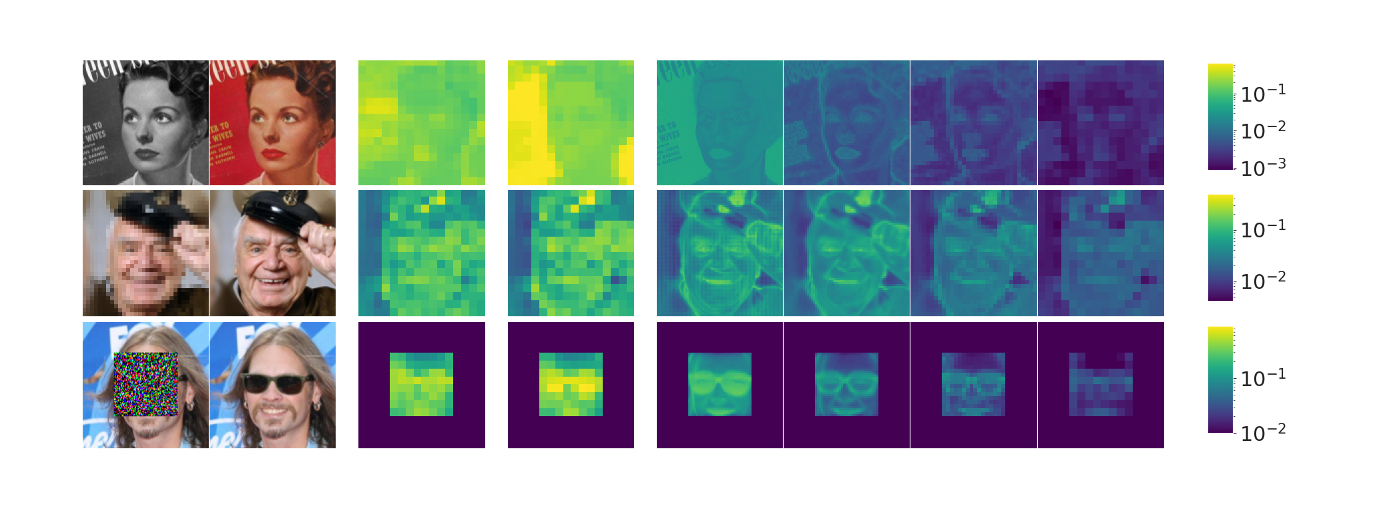}
    \caption{\textbf{Local Experiments:} Uncertainty volume maps for patches applied in image colorization (top), super-resolution (middle), and inpainting (bottom) with E-PUQ, im2im-uq \cite{angelopoulos2022image} and Conffusion \cite{horwitz2022conffusion}. Each pixel in the maps corresponds to the uncertainty volume, defined in Equation~\eqref{eq: uncertainty volume}, of its corresponding patch. These results expose the effectiveness of our method that incorporates spatial correlations, resulting in a reduction of the uncertainty volume.}
    \label{fig: local visualizations}
\end{figure*}

\noindent We apply our proposed methods on RGB patches of increasing size --- 1x1, 2x2, 4x4, and 8x8 --- for image colorization, super-resolution, and inpainting tasks.
The obtained results are illustrated in Figure~\ref{fig: local exact puq results} and Figure~\ref{fig: local adaptive and reduced puq results}, where Figure~\ref{fig: local exact puq results} compares our exact procedure, E-PUQ, to baseline methods, and Figure~\ref{fig: local adaptive and reduced puq results} examines our approximation procedures, DA-PUQ and RDA-PUQ.
In Table~\ref{tab: local results} we present a numerical comparison of uncertainty volumes across tasks at 8x8 patch resolution.
We also provide visual representations of the uncertainty volume maps for patches at varying resolutions in Figure~\ref{fig: local visualizations}.

The results shown in Figure~\ref{fig: local exact puq results} and Figure~\ref{fig: local adaptive and reduced puq results} demonstrate that our method provides smaller uncertainty volumes, and thus more confined uncertainty regions, when compared to previous work in all tasks and patch resolutions, and while satisfying the same statistical guarantees in all cases.
More specifically, Figure~\ref{fig: local exact puq results} compares our exact procedure, E-PUQ, to baseline methods. Following this figure, one can see that using the E-PUQ procedure we obtained an improvement of $\sim \times 100$ in the uncertainty volumes in colorization and an improvement of $\sim \times 10$ in super-resolution and inpainting, when applied to the highest resolution of 8x8.
Additionally, as the patch resolution increases, we observe a desired trend of uncertainty volume reduction, indicating that our method takes into account spatial correlation to reduce uncertainty. Note that even a patch size of $1\times 1$ brings a benefit in the evaluated volume, due to the exploited correlation within the three color channels. E-PUQ reduces trivially to  im2im-uq \cite{angelopoulos2022image} and Conffusion~\cite{horwitz2022conffusion} when applied to scalars ($1 \times 1 \times 1$ patches). 

In Figure~\ref{fig: local adaptive and reduced puq results}, we examine our approximation methods, DA-PUQ and RDA-PUQ, in which we set a relatively small reconstruction risk of $\beta=0.05$. Observe the significantly smaller uncertainty volumes obtained; this effect is summarized in Table~\ref{tab: local results} as well.
Figure~\ref{fig: local adaptive and reduced puq results} also portrays the dimensionality of the uncertainty region used with our method using two overlapping bars. The outer bar in yellow refers to the number of PCs that need to be constructed, denoted as $K$ in DA-PUQ and $\hat{K}$ in RDA-PUQ.
The smaller this number is, the lower the test time computational complexity.
The inner bar in green refers to the average number of the adaptively selected PCs, denoted as $\hat{k}(x)$. A lower value of $\hat{k}(x)$ indicates better interpretability, as fewer PCs are used at inference than those that were constructed.
For example, in the colorization task, it can be seen that the RDA-PUQ procedure is the most computationally efficient methodology, requiring only $\hat{K} \approx 12$ PCs to be constructed at inference, while the DA-PUQ procedure is the most interpretable results, with uncertainty regions consisting of only $\hat{k}(x) \in \{1,2,3\}$ axes.



In all experiments demonstrated in Figure~\ref{fig: local exact puq results} and Figure~\ref{fig: local adaptive and reduced puq results}, it is noticeable that the standard deviation of the interval-size of our approach is higher than that of the baseline methods. This effect happens because a few intervals along the first few PCs are wider than those along the remaining PCs. However, the majority of the interval sizes are significantly smaller, resulting in a much smaller uncertainty volume.
Interestingly, the uncertainty intervals of the DA-PUQ and RDA-PUQ procedures in Figure~\ref{fig: local adaptive and reduced puq results} exhibit larger standard deviation compared to the E-PUQ procedure in Figure~\ref{fig: local exact puq results}. We hypothesize that this is caused when only a few intervals (e.g., 2 intervals) are used for the calibration process while small miscoverage ratio is set by the user ($\alpha = 0.1$). As an example in the case of using 2 intervals with all samples of the calibration set, it is necessary to enlarge all the intervals to ensure the coverage guarantee, resulting in wider intervals over the first few PCs.

The heat maps presented in Figure~\ref{fig: local visualizations} compare the uncertainty volumes of our patch-based E-PUQ procedure to baseline methods. Each pixel in the presented heat maps corresponds to the value of Equation~\eqref{eq: uncertainty volume} evaluated on its corresponding patch. The results show that as the patch resolution increases, pixels with strong correlation structure, such as pixels of the background area, also exhibit lower uncertainty volume in their corresponding patches. This indicates that the proposed method indeed takes into account spatial correlation, leading to reduced uncertainty volume.

\subsection{Global Experiments on Images}

\begin{table}[]
\centering
\caption{\textbf{Global Experiments:} Quantitative comparison of the means and standard deviations of our globally applied PUQ method in the colorization task, utilizing the proposed DA-PUQ (see Section~\ref{sec: Dimensional Adaptive PUQ}) and RDA-PUQ (see Appendix~\ref{app: Reduced Dimensional Adaptive PUQ}) procedures.}
\label{tab: global results}
\resizebox{\columnwidth}{!}{%
\begin{tabular}{@{}lccc@{}}
\toprule
                & \textbf{Recons. Risk} & \textbf{Dim. $\hat{k}(x)$ / $K$} & \textbf{Uncert. Volume} \\ \midrule
im2im-uq \cite{angelopoulos2022image}        & $0$                  & $49152$ / $49152$                    & $\expnumber{1.4}{-1} \pm \expnumber{3.2}{-2}$                    \\
Conffusion \cite{horwitz2022conffusion}      & $0$                  & $49152$ / $49152$                    & $\expnumber{1.4}{-1} \pm \expnumber{5.5}{-2}$                    \\
\textbf{DA-PUQ}  & $\expnumber{5.0}{-2} \pm \expnumber{1.1}{-3}$                  & $2.2 \pm 0.93$ / 100                    & $\expnumber{1.2}{-13} \pm \expnumber{5.0}{-14}$                    \\
\textbf{RDA-PUQ} & $\expnumber{4.3}{-2} \pm \expnumber{2.8}{-3}$                  & $5.5 \pm 4.5$ / $22.3 \pm 10.9$                    & $\expnumber{3.1}{-13} \pm \expnumber{2.4}{-13}$                    \\ \bottomrule
\end{tabular}%
}
\end{table}

\begin{figure}[t]
    \centering
    \includegraphics[width=\columnwidth,trim={4cm 3cm 4cm 0},clip]{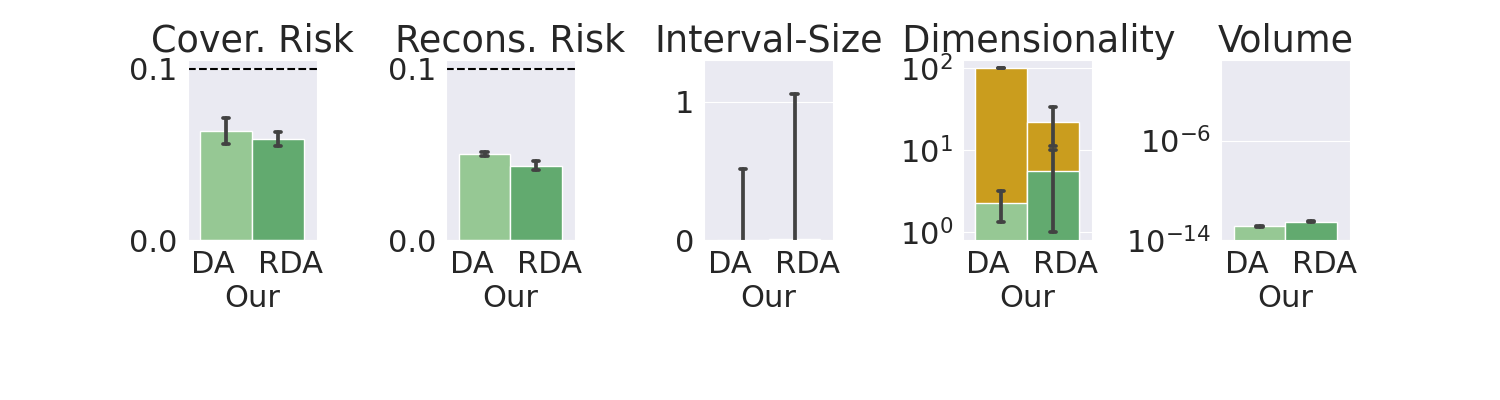}
    \caption{\textbf{Global Experiments:} A comparison of DA-PUQ (see Section~\ref{sec: Dimensional Adaptive PUQ}) and RDA-PUQ (see Appendix~\ref{app: Reduced Dimensional Adaptive PUQ}), when applied globally on the colorization task with $\alpha=\beta=\delta=0.1$ and $q=0.95$.
    The uncertainty volume was computed with $\epsilon = 1e-10$.
    }
    \label{fig: global results}
\end{figure}

\begin{figure}[t]
    \centering
    {\hspace*{0.3cm}\small $x$ ~~~~~~~~~~~~ $y$ ~~~~~~~~~~ Recons. ~~~~~~~ $\hat{v}_1(x)$ ~~~~~~ $\hat{v}_2(x)$ ~~~~~ }\includegraphics[width=\columnwidth,trim={4.2cm 3.3cm 3.5cm 3.3cm},clip]{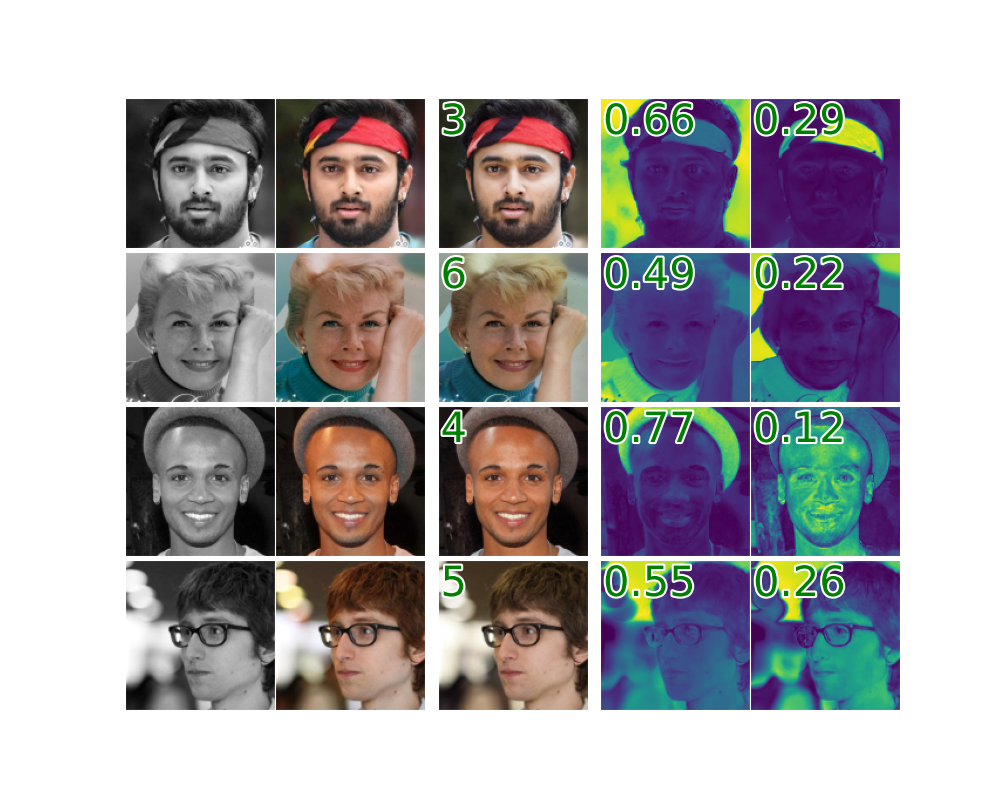}
    \caption{\textbf{Global Experiments:}  Visual presentation of uncertainty regions provided by RDA-PUQ (Appendix~\ref{app: Reduced Dimensional Adaptive PUQ}) when applied globally for the colorization task. The reconstructed image is given by $\hat{\mu}(x)+\sum_{i=1}^{\hat{k}(x)} \hat{v}_i(x)^T y_c\hat{v}_i(x)$, where $y_c := y - \hat{\mu}(x)$. The values of $\hat{k}(x)$, ${\hat w}_1(x)\text{ and }{\hat w}_2(x)$ are shown in the top left corners of the corresponding columns.
    }
    \label{fig: global regions}
\end{figure}

\begin{figure}[t]
    \centering
    {\hspace*{0cm}\small $\mid$------------------ Our -------------------$\mid$ ~~ im2im-uq ~~~ Conffusion}
    \includegraphics[width=\columnwidth,trim={4.2cm 3.3cm 3.5cm 3.3cm},clip]{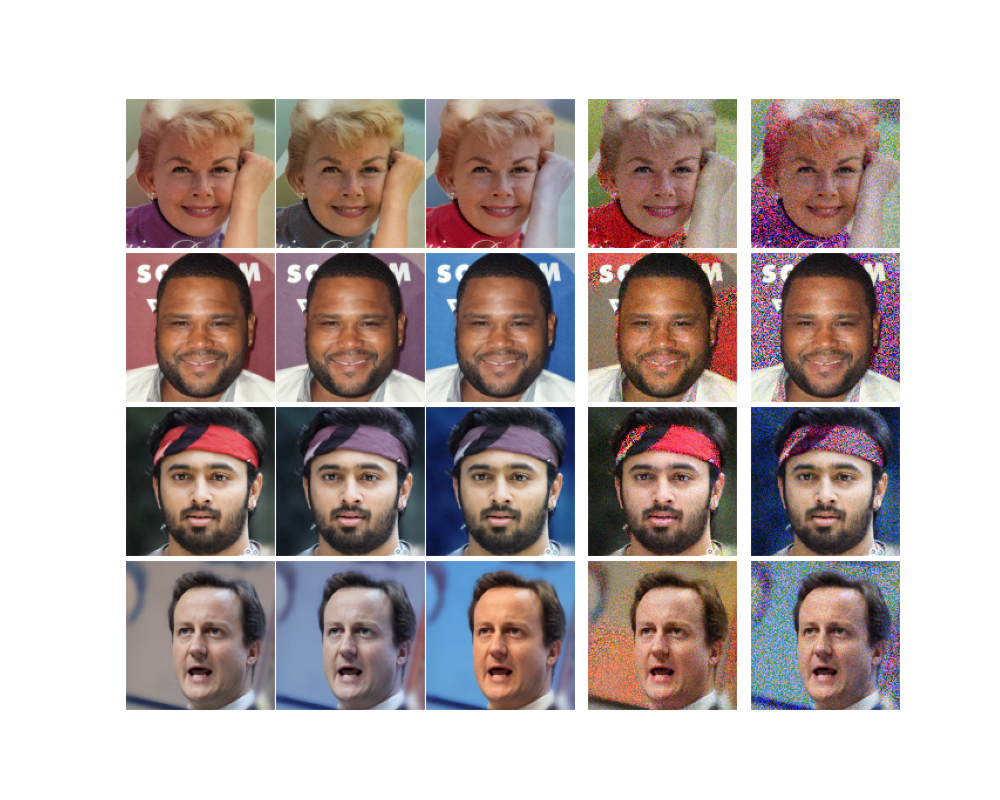}
    \caption{\textbf{Global Experiments:}  Images sampled uniformly from the estimated global uncertainty regions, referring to the colorization task. Using RDA-PUQ results with high-perceptual images, while im2im-uq \cite{angelopoulos2022image} and Conffusion \cite{horwitz2022conffusion} produce unlikely images. These results indicate that our uncertainty regions are significantly more confined than those of previous works.}
    \label{fig: global samples}
\end{figure}


\noindent We turn to examine the effectiveness and validity of DA-PUQ and RDA-PUQ when applied to complete images at a resolution of $128 \times 128$. In this case, the E-PUQ  procedure does not apply, as it requires computing and maintaining $d = 128 \times 128 \times 3$ PCs. We present results for the colorization task hereafter, and refer the reader to Appendix~\ref{app: additional global experiments} for a similar analysis related to super-resolution and inpainting.

While all PUQ procedures can be applied locally for any task, working globally is more realistic in tasks that exhibit strong pixel correlation. Under this setting, most of the image variability could be represented via DA-PUQ or RDA-PUQ while (i) maintaining a small reconstruction risk, and  (ii) using only a few PCs to assess the uncertainty of the entire images. 
We should note that the tasks of super-resolution and inpainting are less-matched to a global mode since they require a larger number of PCs for an effective uncertainty representation -- more on this is discussed in Appendix~\ref{app: additional global experiments}. 


\noindent Figure~\ref{fig: global results} visually demonstrates the performance of our approximation methods, also summarized in Table~\ref{tab: global results}.
These results 
demonstrate that our method provides significantly smaller uncertainty volumes compared to our local results in Figure~\ref{fig: local adaptive and reduced puq results} and previous works, but this comes at the cost of introducing a small reconstruction risk of up to $\beta = 0.1$. 
Observe how our approximation methods improve interpretability: the uncertainty regions consist of only 2-5 PCs in the full dimensional space of the images.
The DA-PUQ procedure produces the tightest uncertainty regions; see the uncertainty volumes in Table~\ref{tab: global results}. In addition, the mean interval-size with our procedures is very small and almost equal to zero, indicating that the constructed uncertainty regions are tight and narrow due to strong correlation structure of pixels. However, similar to the previous results, the standard deviation of interval-size is spread across a wide range. This is because few of the first PCs have wide intervals.
The RDA-PUQ procedure is the most computationally efficient as it required to construct only $\sim$30 PCs during inference to ensure statistical validity.

Figure~\ref{fig: global regions} presents selected uncertainty regions that were provided by our proposed RDA-PUQ procedure when applied globally. As can be seen, the projected ground-truth images using only ${\hat k}(x)$ PCs results in images that are very close to the originals. This indicates that the uncertainty region can describe the spread and variability among solutions with small reconstruction errors. The first two axes of our uncertainty regions exhibit semantic content, which is consistent with a method that accounts for spatial pixel correlation. The fact that these PCs capture foreground/background or full-object content highlights a unique strength of our approach. 
We provide the importance weights of the first two PCs, indicating impressive proportions of variability among projected samples onto these components (see Section~\ref{sec: our implementation}). For example, in the third row, we observe that $77\%$ of the variability in $\thP$ is captured by $\hat{v}_1(x)$, which mostly controls a linear color range of the pixels associated with the hat in the image.
In Figure~\ref{fig: global samples} we visually compare samples that were generated from the corresponding estimated uncertainty regions, by sampling uniformly a high dimensional point (i.e., an image) from the corresponding hyper-rectangle.
\Review{For further details regarding this study, please refer to Appendix~\ref{app: samples from uncertainty region}.}
As can be seen, the samples extracted from our uncertainty region are of high perceptual quality, whereas im2im-uq \cite{angelopoulos2022image} and Conffusion \cite{horwitz2022conffusion} produce highly improbable images. This testifies to the fact that our method provides much tighter uncertainty regions, whereas previous work results in exaggerated regions that contain unlikely images.
In addition to the above, we present in Appendix~\ref{app: corners} a visualization of the lower and upper \emph{corners} of the uncertainty regions produced by our method, comparing them to those produced by previous work \cite{angelopoulos2022image, horwitz2022conffusion}.


\section{Concluding Remarks}

\noindent This paper presents ``Principal Uncertainty Quantification'' (PUQ), a novel and effective approach for quantifying uncertainty in any image-to-image task. PUQ takes into account the spatial dependencies between pixels in order to achieve significantly tighter uncertainty regions. 
The experimental results demonstrate that PUQ outperforms existing methods in image colorization, super-resolution and inpainting, by improving the uncertainty volume. Additionally, by allowing for a small reconstruction error when recovering ground-truth images, PUQ produces tight uncertainty regions with a few axes and thus improves computational complexity and interpretability at inference. As a result, PUQ achieves state-of-the-art performance in uncertainty quantification for image-to-image problems.

Referring to future research, more sophisticated choices that rely on recent advancements in stochastic image regression models could be explored, so as to improve the complexity of our proposed approximation phase. Further investigation into alternative geometries for uncertainty regions could be interesting in order to reduce the gap between the provided region of uncertainty and the high-density areas of the true posterior distribution. This includes an option to divide the spatial domain into meaningful segments, while minimizing the uncertainty volume, or consider a mixture of Gaussians modeling of the samples of the estimated posterior distribution.
\Review{Additionally, exploring alternative diffusion models and various conditional stochastic samplers presents an interesting path for future investigation. This could involve comparing different conditional samplers, potentially offering an alternative approach to the utilization of FID scores.}



\bibliographystyle{unsrt}
\bibliography{references}


%

\appendix

\subsection{Visualizing the Principal Component Vectors}
\label{app: pcs motivation}

\begin{figure*}[t]
    \centering
    \includegraphics[width=0.9\textwidth,trim={1cm 1.4cm 0 0.2cm},clip]{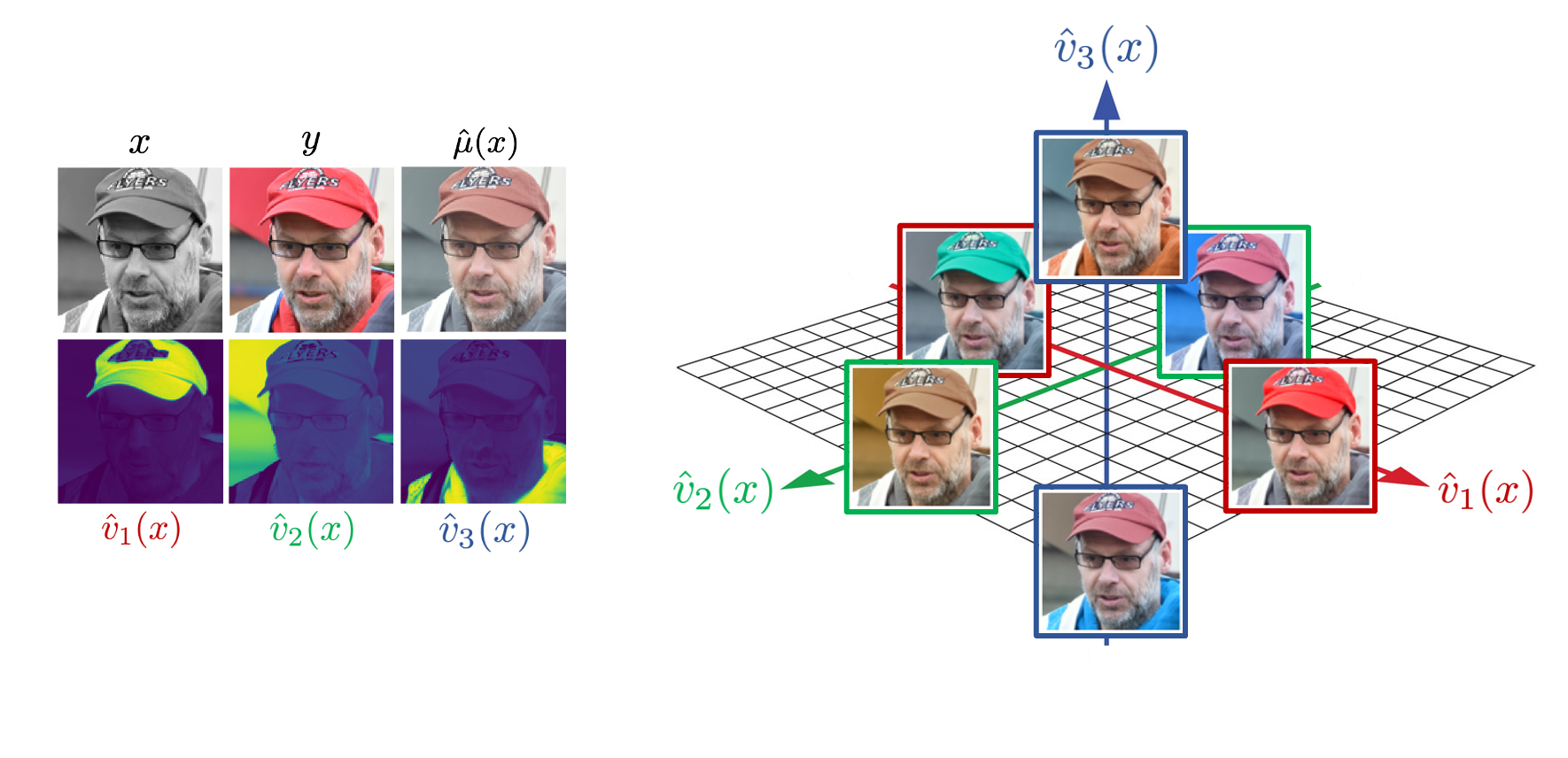}
    \caption{A visual representation demonstrating the intuition behind utilizing principal components (PCs) as the basis, $\hat{B}(x)$, in Equation~\eqref{eq: intervals} for the colorization task. The left part illustrates that the PCs incorporate spatial correlation, with $\hat{v}_1(x)$ primarily controlling the hat color, $\hat{v}_2(x)$ governing the background color, and $\hat{v}_3(x)$ influencing the clothing color. On the right side, an illustration of the uncertainty region is presented, composed of these axes, where the origin is $\hat{\mu}(x)$, and each image is defined by $\hat{\mu}(x) + \hat{v}_i(x)^T y_c + a$, where $y_c := y - \hat{\mu}(x)$, and $a\in\dR$ is a controllable parameter that moves along the axis.}
    \label{fig: pcs motivation}
\end{figure*}

\noindent Figure~\ref{fig: pcs motivation} depicts the role of the Principal Component (PCs) vectors in the context of the image colorization task. This figure provides an intuition behind employing these vectors for the uncertainty quantification. We show the estimation of the first three PCs using our globally applied PUQ and visualize the uncertainty region formed by these axes. 
Our approach facilitates efficient exploration within the uncertainty region, thanks to the linear axes that incorporate spatial correlation, as illustrated by the visualization of $\hat{v}_1(x)$, $\hat{v}_2(x)$, and $\hat{v}_3(x)$.

\subsection{Coverage Loss Justification}
\label{app: coverage loss justification}

\noindent This section aims to justify our choice for the loss-function for tuning $\lambda$ in Equation~\eqref{eq: full coverage loss}, and the weights used in it, $\hat{w}_i(x)$. Recall, this expression is given as:
{\small\begin{equation*}
\mathcal{L} (x, y; \lambda) := \sum_{i=1}^{d} \hat{w}_i(x) \cdot \mathbf{1}\left\{\hat{v}_i(x)^T y \not\in \mathcal{T}_\lambda(x; \hat{B}(x))_i \right\}. 
\end{equation*}}

Our starting point is the given $d$-dimensional hyper-rectangle obtained from the approximation phase, oriented along the $d$ PC directions. This shape serves as our initially estimated uncertainty region. Given the calibration data, $\mathcal{S}_\text{cal} := \{ ( x_i , y_i ) \}_{i=1}^n$, our goal is to inflate (or deflate, if this body proves to be exaggerated) this shape uniformly across all axes so that it contains the majority of the ground truth examples. 

Focusing on a single pair from this dataset, $(x,y)$, the degraded image $x$ is used to ignite the whole approximation phase, while the ground truth $y$ serves for assessing the obtained hyper-rectangle, by considering the projected coordinates $\{\hat{v}_i(x)^T y_c\}_{i=1}^d$, where $y_c := y - \hat{\mu}(x)$. The following function measures a potential deviation in the $i$-th axis,
{\small\begin{align}
\label{eq: potential deviation coverage}
    h_i(x, y) :=& \max \bigl\{ \hat{v}_i(x)^T y_c - \hat{u}(x)_i, 0 \bigr\} \\
    \nonumber &+ \max \bigl\{ -\hat{v}_i(x)^T y_c + \hat{l}(x)_i , 0 \bigr\}.
\end{align} } 
Written differently, this expression is also given by
{\small\begin{equation*}
    h_i(x, y) := \begin{cases}
        \hat{v}_i(x)^T y_c - \hat{u}(x)_i & \text{if } \hat{v}_i(x)^T y_c > \hat{u}(x)_i>0 \\
        \hat{l}(x)_i - \hat{v}_i(x)^T y_c & \text{if } \hat{v}_i(x)^T y_c < -\hat{l}(x)_i<0 \\
        0 & \text{otherwise}.
    \end{cases}
\end{equation*}}
If positive, this implies that in this axis the example spills outside the range of the rectangle, and the value itself is the distance from it's border. 

The following expression quantifies the weighted amount of energy that should be invested in projecting back the $\{\hat{v}_i(x)^T y\}_{i=1}^d$ coordinates to the closest border point: 
{\small\begin{align}
\label{eq: coverage energy}
    \text{Energy}(x,y) = \sum_{i=1}^d \hat{\sigma}_i(x)^2 h_i(x, y)^2.
\end{align}}
Note that in our weighting we prioritize high-variance axes, in which deviation from the boundaries is of greater impact. Naturally, we should tune $\lambda$, which scales $\hat{u}(x)_i $ and $\hat{l}(x)_i $, so as to reduce this energy below a pre-chosen threshold, thus guaranteeing that the majority of ground truth images fall within the hyper-rectangle. While this expression is workable, it suffers from two shortcomings: (1) It is somewhat involved to compute; and (2) The threshold to use with it is hard to interpret and thus to choose. Therefore, similar to previous approaches \cite{angelopoulos2022image, horwitz2022conffusion}, we opted in this work for a binary version of Equation~\eqref{eq: potential deviation coverage} of the form
{\small\begin{equation}
    b_i(x, y) := \begin{cases}
        1 & \text{if } h_i(x, y) > 0 \\
        0 & \text{otherwise}.
    \end{cases}
\end{equation}}
In addition, we divide the energy expression, defined in Equation~\eqref{eq: coverage energy}, by the sum of squares of all the singular values, and this way obtain exactly $\mathcal{L} (x, y; \lambda)$ as in Equation~\eqref{eq: full coverage loss}. Observe that, by definition, we get that $0\le \mathcal{L} (x, y; \lambda) \le 1$, where the bottom bound corresponds to a point fully within the rectangle, and the upper bound for the case where the point is fully outside in all axes. Therefore, thresholding the expectation of this value with $\alpha \ll 1$ is intuitive and meaningful. 

\subsection{Reconstruction Loss Justification}
\label{app: reconstruction loss justification}

\noindent This section aims to discuss our choice for the loss function for tuning $\lambda_1$ in Equation~\eqref{eq: interpertable reconstruction loss}, given by
{\small\begin{align*}
\mathcal{L}_1(x,y;\lambda_1) := \hat{\mathcal{Q}}_q \left( \left\{ \left| \sum_{j=1}^{\hat{k}(x;\lambda_1)} \hat{v}_j(x)^T y_c \hat{v}_j(x) - y_c \right|_i \right\}_{i=1}^d \right) ~.
\end{align*}}
Recall the process: We begin with  $K \leq d$ PCs obtained from the approximation phase, 
and then choose $\hat{k}(x;\lambda_1) \leq K$ of them as instance-specific number of PCs 
for the evaluation of the uncertainty. 

Given a calibration 
pair $(x,y)$, 
$x$ is used to derive $\hat{k}(x;\lambda_1)$, defining a low-dimensional subspace $\hat{V}(x) := [\hat{v}_1(x),\dots,\hat{v}_{\hat{k}(x;\lambda_1)}(x)] \in \dR^{\hat{k}(x;\lambda_1) \times d}$. This, along with the conditional-mean, $\hat{\mu}(x)$, represent $\tP$ as an affine subspace. The ground-truth image $y$ is then projected onto this slab via:
{\small\begin{align}
\text{Projection}(y) &:= \hat{\mu}(x) + \hat{V}(x) \hat{V}(x)^T y_c \\
\nonumber &= \hat{\mu}(x) + \sum_{j=1}^{\hat{k}(x;\lambda_1)} \hat{v}_j(x)^T y_c \hat{v}_j(x) ~,
\end{align}}
where $y_c := y - \hat{\mu}(x)$.

The parameter $\lambda_1$ should be tuned so as to guarantee that this projection entails a bounded error, $dist(y,\text{Projection}(y))$ in expectation. A natural distance measure to use here is the $L_2$-norm of the difference, which aligns well with our choice to use SVD in the approximation phase. However, $L_2$ accumulates the error over the whole support, thus losing local interpretability. An alternative is using $L_\infty$ which quantifies the worst possible pixelwise error induced by the low-dimensional projection, 
{\small\begin{align}
\nonumber
dist(y,\text{Projection}(y)) &:= \left\| \sum_{j=1}^{\hat{k}(x;\lambda_1)} \hat{v}_j(x)^T y_c \hat{v}_j(x) - y_c \right\|_{\infty} ~.\end{align}}
While this measure is applicable in many tasks, there are cases (e.g., inpainting) in which controlling a small maximum error requires the use of a large number of PCs, $\hat{k}(x;\lambda_1)$.
To address this, we propose a modification 
by considering the maximum error over a user-defined ratio of pixels, $q \in (0,1)$, a value close to $1$.
This is equivalent to determining the $q$-th empirical quantile, $\hat{\mathcal{Q}}_q$, of the error values among the pixels, providing a more flexible and adaptive approach, which also aligns well with the rationale of uncertainty quantification, in which the statistical guarantees are given with probabilistic restrictions.

\subsection{Reduced Dimension-Adaptive PUQ}
\label{app: Reduced Dimensional Adaptive PUQ}

\noindent The DA-PUQ procedure (see Section~\ref{sec: Dimensional Adaptive PUQ}) reduces the number of PCs to be constructed to $K \leq d$ while using $\hat{k}(x;\hat{\lambda}_1) \leq K$ PCs, leading to increased efficiency in both time and space during inference.
However, determining manually the smallest $K$ value that can guarantee both Equation~\eqref{eq: theoretical coverage gaurentee} and Equation~\eqref{eq: theoretical reconstruction gaurentee} with high probability can be challenging.
To address this, we propose an expansion of the DA-PUQ procedure; the \emph{Reduced Dimension-Adaptive PUQ} (RDA-PUQ) procedure that also controls the maximum number of PCs required for the uncertainty assessment.
While this approach is computationally intensive during calibration,
it is advantageous for inference as it reduces the number of samples required to construct the PCs using Algorithm~\ref{alg: approximation phase}.

Specifically, for each input instance $x$ and its corresponding ground-truth value $y$ in the calibration data, we use the estimators obtained in the approximation phase, to estimate $\hat{K}_{\lambda_3}$ PCs of possible solutions, denoted by $\hat{B}(x)$, their corresponding importance weights, denoted by $\hat{w}(x)$, the conditional mean denoted by $\hat{\mu}(x)$, and the lower and upper bounds denoted by $\tilde{l}(x)$ and $\tilde{u}(x)$, respectively.
Note that these estimates are now depend on $\lambda_3$, we omit the additional notation for simplicity.
Then, for each choice of $\lambda_3$, we use these $\hat{K}_{\lambda_3}$-dimensional estimates exactly as in the DA-PUQ procedure to achieve both the coverage and reconstruction guarantees of Equation~\eqref{eq: theoretical coverage gaurentee} and Equation~\eqref{eq: theoretical reconstruction gaurentee} with high probability.


Similar to previous approaches, we aim to minimize the uncertainty volume, defined in Equation~\eqref{eq: uncertainty volume}, for the scaled $\hat{K}_{\lambda_3}$-dimensional intervals where any additional axis ($d - \hat{K}_{\lambda_3}$ axes) is fixed to zero. We denote the uncertainty volume in this setting as $\mathcal{V}_{\lambda_1,\lambda_2,\lambda_3}$.
The minimization of $\mathcal{V}_{\lambda_1,\lambda_2,\lambda_3}$ is achieved by minimizing $\lambda_1$, $\lambda_2$ and $\lambda_3$, while ensuring that the guarantees of Equation~\eqref{eq: theoretical coverage gaurentee} and Equation~\eqref{eq: theoretical reconstruction gaurentee} are satisfied with high probability.
This can be provided using a conformal prediction scheme, for example, through the LTT \cite{angelopoulos2021learn} calibration scheme, which ensures that the following holds:
{\small\begin{align}
\label{eq: Reduced Dimensional Adaptive PUQ guarantee}
\dP \left( 
\begin{array}{c}
\dE[\mathcal{L}_1 (x, y; \hat{\lambda}_1, \hat{\lambda}_3)] \leq \beta \\ \dE[\mathcal{L}_2 (x, y; \hat{\lambda}_1, \hat{\lambda}_2, \hat{\lambda}_3)] \leq \alpha 
\end{array}
\right) \geq 1 - \delta ~,
\end{align}}

where $\hat{\lambda}_1$, $\hat{\lambda}_2$ and $\hat{\lambda}_3$ are the minimizers for the uncertainty volume among valid calibration parameter results, $\hat{\Lambda}$, obtained through the LTT procedure.
Note that the loss functions, $\mathcal{L}_1$ and $\mathcal{L}_2$, in the above are exactly those of the DA-PUQ procedure, defined in Equation~\eqref{eq: interpertable reconstruction loss} and Equation~\eqref{eq: dimensional adaptive loss}, while replacing $K$ with $\hat{K}_{\lambda_3}$.

Intuitively, Equation~\eqref{eq: Reduced Dimensional Adaptive PUQ guarantee} guarantees that a fraction $q$ of the ground-truth pixel values is recovered with an error no greater than $\beta$ using no more than $\hat{K}_{\hat{\lambda}_3}$ principal components, and a fraction of more than $1-\alpha$ of the projected ground-truth values onto the first $\hat{k}(x;\hat{\lambda}_1)$ principal components (out of $\hat{K}_{\hat{\lambda}_3})$ are contained in the uncertainty intervals, with a probability of at least $1-\delta$.
The RDA-PUQ procedure is formally described in Algorithm~\ref{alg: reduced dimensional adaptive puq}.

\begin{algorithm}[htb]
\small
\caption{Reduced Dimension-Adaptive PUQ Proc.}
\label{alg: reduced dimensional adaptive puq}
\begin{algorithmic}[1]

\Require{Calibration set $\mathcal{S}_{\text{cal}} := \{ x_i , y_i \}_{i=1}^n$.
Scanned calibration parameter values $\Lambda^1 \gets [ 1 \dots {\lambda_1}_{\text{max}} ]$, $\Lambda^2 \gets [ 1 \dots {\lambda_2}_{\text{max}} ]$ and $\Lambda^3 \gets [ 1 \dots {\lambda_3}_{\text{max}} ]$.
Maximal PCs number $K \leq d$. Approximation phase estimators $\hat{B},\hat{w},\hat{\mu},\tilde{u},\tilde{l}$. Recovered pixels ratio $q \in (0,1)$. Reconstruction error $\beta \in (0,1)$. Misscoverage ratio $\alpha \in (0,1)$. Calibration error level $\delta \in (0,1)$.}


\For{$(x,y) \in \mathcal{S}_\text{cal}$}

\For{$\lambda_3 \in \Lambda_3$}

\Comment{Reduce dimensionality}

\State $\hat{K}_{\lambda_3} \gets \lfloor K \cdot \lambda_3 \rfloor$

\State $\hat{B}(x),\hat{w}(x),\hat{\mu}(x),\tilde{u}(x),\tilde{l}(x) \gets$
Apply Algorithm~\ref{alg: approximation phase} \par\hspace*{3em} to $x$, with the choice of $\hat{K}_{\lambda_3}$ samples

\For{$\lambda_1 \in \Lambda_1$}

\Comment{Compute adaptive dimensionality, Equation~\eqref{eq: adaptive k}}

\State $\hat{k}(x;\lambda_1,\lambda_3) \gets $\par\hspace*{4em}$ \min_k \left\{ k : \sum_{i=1}^{\hat{K}_{\lambda_3}} \hat{w}_i(x) \geq \lambda_1 \right\} $

\Comment{Compute reconstruction loss, Equation~\eqref{eq: interpertable reconstruction loss}}

\State $y_c \gets y - \hat{\mu}(x)$

\State $\mathcal{L}_1(x,y;\lambda_1,\lambda_3) \gets $\par
\hspace*{-1em}$\hat{\mathcal{Q}}_q \bigg( \bigg\{ \bigg| \sum_{j=1}^{\hat{k}(x;\lambda_1,\lambda_3)} \hat{v}_j(x)^T y_c \hat{v}_j(x) - y_c \bigg|_i \bigg\}_{i=1}^d \bigg)$

\For{$\lambda_2 \in \Lambda_2$}

\Comment{Scale uncertainty intervals}

\State \smash{$\hat{u}(x) \gets \lambda_2 \tilde{u}(x)$ and $\hat{l}(x) \gets \lambda_2 \tilde{l}(x)$}

\State $\mathcal{T}_{\lambda_2}(x; \hat{B}(x)) \gets$
\par
\hspace*{6em} Equation~\eqref{eq: intervals} using $\hat{\mu}(x),\hat{u}(x),\hat{l}(x)$

\Comment{Compute weighted coverage loss, Equation~\eqref{eq: full coverage loss}}

\State \smash{$\mathcal{L}_2(x,y;\lambda_1,\lambda_2,\lambda_3) \gets \sum_{i=1}^{\hat{k}(x;\lambda_1,\lambda_3)} \hat{w}_i(x) \cdot $}\par
\hspace*{6em}\smash{$ \mathbf{1}\left\{{\hat{v}_i(x)}^T y \not\in \mathcal{T}_{\lambda_2}(x; \hat{B}(x))_i \right\}$}

\EndFor

\EndFor

\EndFor

\EndFor

\State $\hat{\Lambda} \gets$ Extract valid $\lambda$s from LTT \cite{angelopoulos2021learn} applied on $\{ ( \mathcal{L}_1(x,y;\lambda_1,\lambda_3) , \mathcal{L}_2(x,y;\lambda_1,\lambda_2,\lambda_3) ) : {(x,y) \in S_{\text{cal}}, \lambda_1 \in \Lambda^1 , \lambda_2 \in \Lambda^2 , \lambda_3 \in \Lambda^3}\}$ at risk levels ($\beta,\alpha$).

\Comment{Compute the minimizers for the uncer. volume, Equation~\eqref{eq: uncertainty volume}}

\State $\hat{\lambda}_1, \hat{\lambda}_2, \hat{\lambda}_3 \gets \arg \min_{\lambda_1,\lambda_2,\lambda_3 \in \hat{\Lambda}} \left\{ \frac{1}{n} \sum_{i=1}^n \mathcal{V}_{\lambda_1,\lambda_2}(x_i;\hat{B}(x_i)) \right\}$

\Ensure{Given a new instance $x \in \mathcal{X}$, obtain valid uncertainty intervals for it,  $\mathcal{T}_{\hat{\lambda}_2}(x; \hat{B}(x))$ over $\hat{k}(x;\hat{\lambda}_1) \leq \hat{K}_{\hat{\lambda}_3}$ PCs.}

\end{algorithmic}
\end{algorithm}

\subsection{Experimental Details}
\label{app: Experimental Details}

\noindent This section provides details of the experimental methodology employed in this study, including the datasets used, architectures implemented, and the procedural details and hyperparameters of our method.

\subsubsection{Datasets and Preprocessing}

Our machine learning system was trained using the Flickr-Faces-HQ (FFHQ) dataset \cite{karras2019style}, which includes 70,000 face images at a resolution of 128x128.
We conducted calibration and testing on the CelebA-HQ (CelebA) dataset \cite{karras2017progressive}, which also consists of face images and was resized to match the resolution of our training data.
To this end, we randomly selected 2,000 instances from CelebA, of which 1,000 were used for calibration and 1,000 for testing.
For the colorization experiments, a grey-scale transformation was applied to the input images.
For the super-resolution experiments, patches at a resolution of 32x32 were averaged to reduce the input image resolution by a factor of 4 in each dimension.
For the inpainting experiments, we randomly cropped pixels from the input images during the training phase, either in squares or irregular shapes; while for the calibration and testing data, we cropped patches at a resolution of 64x64 at the center of the image.

\subsubsection{Architecture and Training}

In all our experiments, we applied the approximation phase using recent advancements in conditional image generation through diffusion-based models, while our proposed general scheme in Algorithm~\ref{alg: general scheme} can accommodate any stochastic regression solvers for inverse problems, such as conditional GANs \cite{mirza2014conditional}.
In all tasks, we utilized the framework for conditional diffusion-based models proposed in the SR3 work \cite{saharia2022image}, using a U-Net architecture.
For each of the three tasks, we trained a diffusion model separately and followed the training regimen outlined in the code of \cite{saharia2022image}.
To ensure a valid comparison with the baseline methods, we implemented them using the same architecture and applied the same training regimen.
All experiments, including the baseline methods, were trained for 10,000 epochs with a batch size of 1,024 input images.

\subsubsection{PUQ Procedures and Hyperparameters}

Our experimental approach follows the general scheme presented in Algorithm~\ref{alg: general scheme} and consists of 2 sets of experiments: local experiments on patches and global experiments on entire images.
For the local experiments, we conducted 4 experiments of the E-PUQ procedure (detailed in Section~\ref{sec: Exact PUQ}) on RGB patch resolutions of 1x1, 2x2, 4x4, and 8x8. We used $K=3$, $K=12$, $K=48$, and $K=192$ PCs for each resolution, respectively.
We set $\alpha=\delta=0.1$ to be the user-specified parameters of the guarantee, defined in Equation~\eqref{eq: Exact PUQ guarantee}.
In addition, we conducted another 2 experiments of the DA-PUQ (detailed in Section~\ref{sec: Dimensional Adaptive PUQ}), and RDA-PUQ (detailed in Appendix~\ref{app: Reduced Dimensional Adaptive PUQ}) procedures on RGB patch resolution of 8x8.
We set $q=0.9$, $\beta = 0.05$ and $\alpha=\delta=0.1$, to be the user-specified parameters of the guarantees of both Equation~\eqref{eq: Dimensional Adaptive PUQ guarantee} and Equation~\eqref{eq: Reduced Dimensional Adaptive PUQ guarantee}.
In total, we conducted $18$ local experiments across three tasks.
For the global experiments, we used entire images at a resolution of 128x128, in which we applied the DA-PUQ and the RDA-PUQ procedures.
As global working is suitable for tasks that exhibit strong pixel correlation, we applied these experiment only on the task of image colorization.
We set $q=0.95$, $\beta=\alpha=\delta=0.1$, to be the user-specified parameters of the guarantees of both Equation~\eqref{eq: Dimensional Adaptive PUQ guarantee} and Equation~\eqref{eq: Reduced Dimensional Adaptive PUQ guarantee}.
Both locally and globally, for the DA-PUQ and RDA-PUQ experiments, we used $K=100$ PCs in the colorization task and $K=200$ PCs in super-resolution and inpainting.
We note that in the RDA-PUQ experiments, we used $\hat{K}$ PCs during inference, as discussed in Appendix~\ref{app: Reduced Dimensional Adaptive PUQ}.
In all experiments we used $\epsilon = 1e-10$ for the computation of the uncertainty volume, defined in Equation~\eqref{eq: uncertainty volume}.

\subsection{\Review{Comparative Samples from Uncertainty Regions}}
\label{app: samples from uncertainty region}

\Review{\noindent We provide here more details referring to the experiment involving a visualization of samples drawn from the uncertainty regions of baseline methods \cite{angelopoulos2022image,horwitz2022conffusion} and our proposed approach. We note that the baseline methods lack such an experiment.

This experiment was conducted across entire images, showing that our uncertainty region is much tighter, containing highly probable image candidates, compared to the pixelwise baseline methods. 
These methods tend to generate exaggerated uncertainty regions that encompass a range of noisy images, diverging from the posterior distribution of images given a measurement.
Our success in producing more confined regions, encompassing the ground truth within them, is a direct consequence of the incorporation of spatial correlations.

To justify this claim, we trained the identical architecture for each baseline method and applied the same training regime that was utilized in our approach, leveraging the official code of both methods.
Each baseline method generates uncertainty intervals via pixel-based uncertainty maps, which is equivalent to our general definition of uncertainty intervals defined by Equation~\eqref{eq: intervals}, while employing standard basis vectors.
Therefore, we uniformly sampled values within the uncertainty intervals of each approach, including our own, and showcased the resulting images.}

\subsection{Additional Global Experiments}
\label{app: additional global experiments}

\begin{figure}[t]
    \centering
    \includegraphics[width=\columnwidth,trim={0cm 1cm 1cm 0},clip]{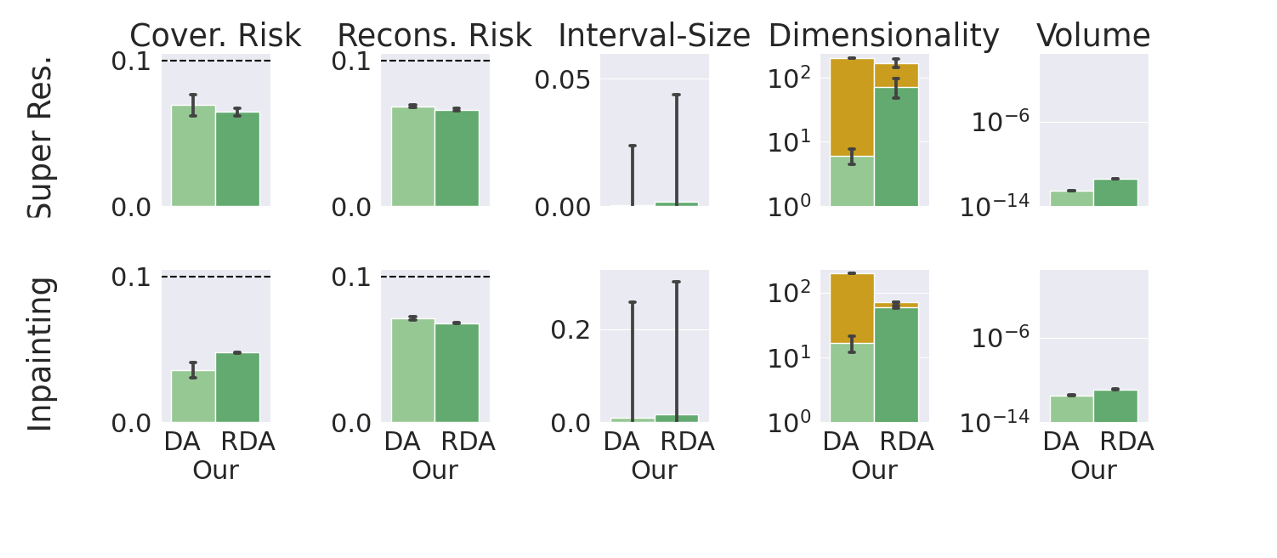}
    \caption{\textbf{Global Experiments:} A comparison of DA-PUQ (see Section~\ref{sec: Dimensional Adaptive PUQ}) and RDA-PUQ (see Appendix~\ref{app: Reduced Dimensional Adaptive PUQ}), when applied globally on super-resolution and inpainting tasks with $\alpha=\beta=\delta=0.1$, where in super-resolution we set $q=0.95$ and in inpainting we set $q=0.8$.
    The uncertainty volume was computed with $\epsilon = 1e-10$.
    }
    \label{fig: global results sr inp}
\end{figure}

\begin{figure}[t]
    \centering
    {\hspace*{0.3cm}\small $x$ ~~~~~~~~~~~~ $y$ ~~~~~~~~~~ Recons. ~~~~~~~ $\hat{v}_1(x)$ ~~~~~~ $\hat{v}_2(x)$ ~~~~~ }\includegraphics[width=\columnwidth,trim={4.2cm 3.3cm 3.5cm 3.3cm},clip]{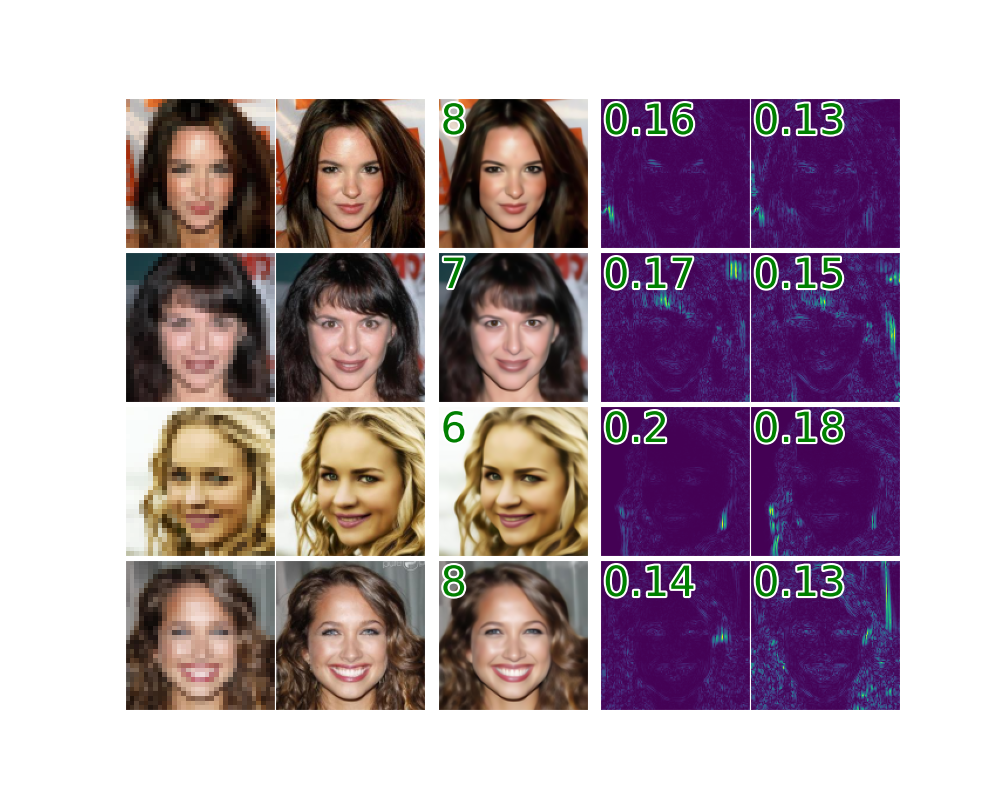}
    \caption{\textbf{Global Experiments:}  Visual presentation of uncertainty regions provided by DA-PUQ when applied globally for the super-resolution task. The reconstructed image is given by $\hat{\mu}(x)+\sum_{i=1}^{\hat{k}(x)} \hat{v}_i(x)^T y_c\hat{v}_i(x)$, where $y_c := y - \hat{\mu}(x)$. The values of $\hat{k}(x)$, ${\hat w}_1(x)\text{ and }{\hat w}_2(x)$ are shown in the top left corners of the corresponding columns.
    }
    \label{fig: global regions sr}
\end{figure}

\begin{figure}[t]
    \centering
    {\hspace*{0cm}\small $\mid$------------------ Our -------------------$\mid$ ~~ im2im-uq ~~~ Conffusion}
    \includegraphics[width=\columnwidth,trim={4.2cm 3.3cm 3.5cm 3.3cm},clip]{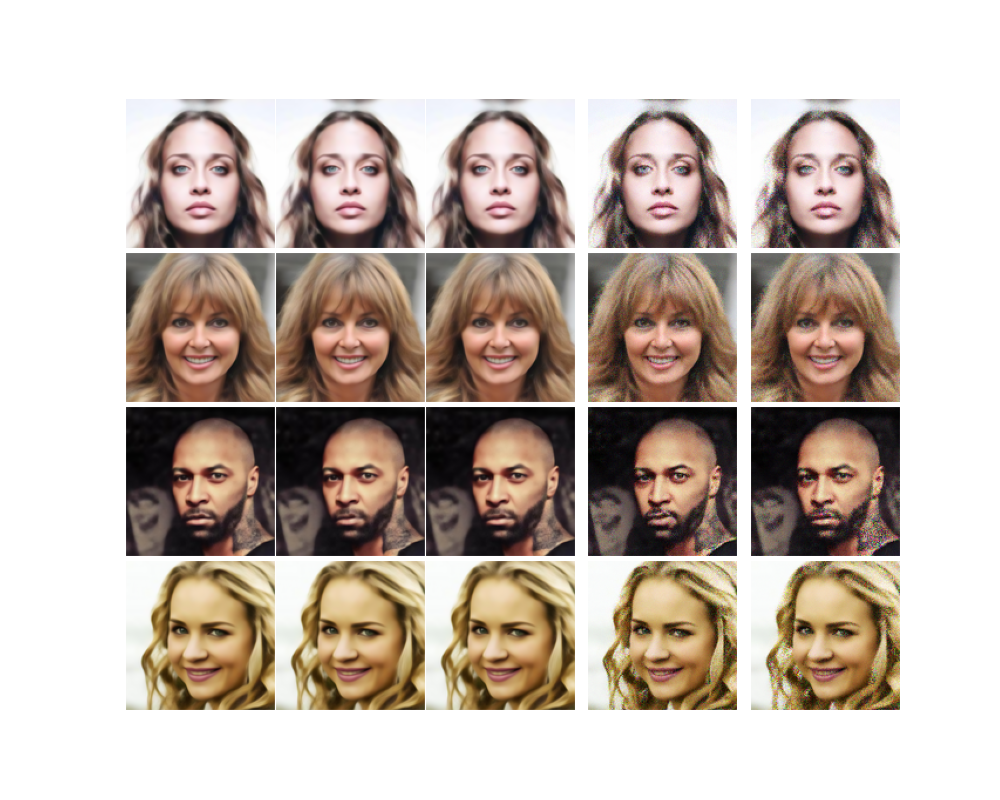}
    \caption{\textbf{Global Experiments:}  Images sampled uniformly from the estimated global uncertainty regions, referring to the super-resolution task. Using DA-PUQ results with high-perceptual images, while im2im-uq \cite{angelopoulos2022image} and Conffusion \cite{horwitz2022conffusion} produce unlikely images. These results indicate that our uncertainty regions are significantly more confined than those of previous works.}
    \label{fig: global samples sr}
\end{figure}

\begin{figure}[t]
    \centering
    {\hspace*{0.3cm}\small $x$ ~~~~~~~~~~~~ $y$ ~~~~~~~~~~ Recons. ~~~~~~~ $\hat{v}_1(x)$ ~~~~~~ $\hat{v}_2(x)$ ~~~~~ }\includegraphics[width=\columnwidth,trim={4.2cm 3.3cm 3.5cm 3.3cm},clip]{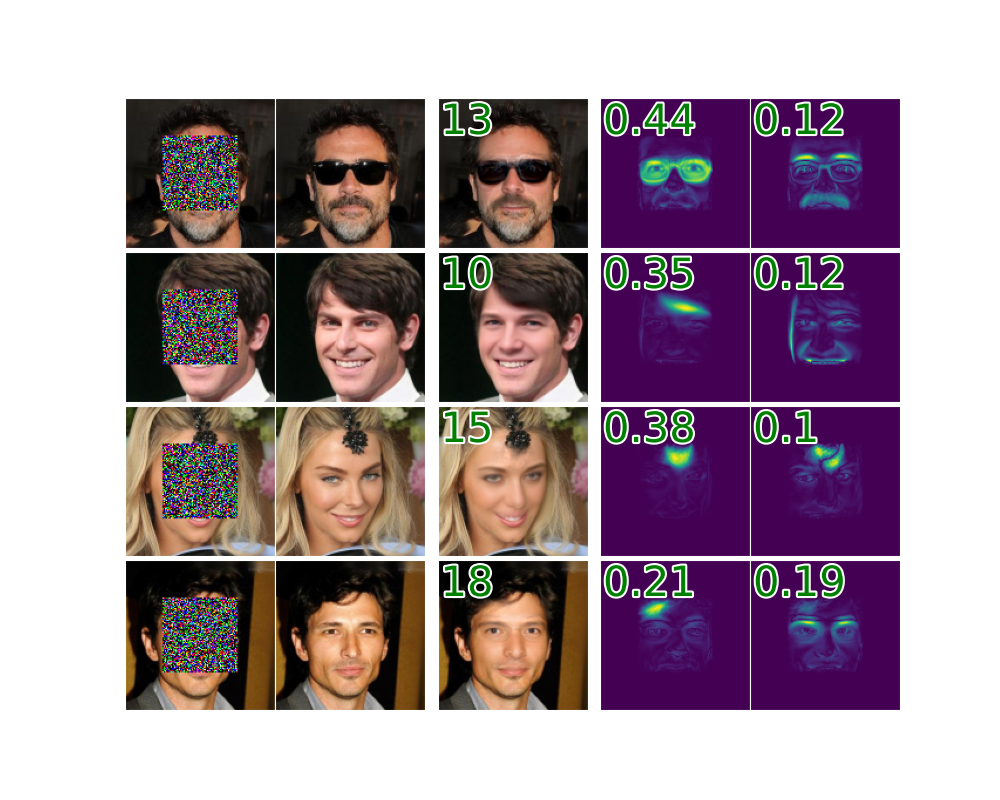}
    \caption{\textbf{Global Experiments:}  Visual presentation of uncertainty regions provided by DA-PUQ when applied globally for the inpainting task. The reconstructed image is given by $\hat{\mu}(x)+\sum_{i=1}^{\hat{k}(x)} \hat{v}_i(x)^T y_c\hat{v}_i(x)$, where $y_c := y - \hat{\mu}(x)$. The values of $\hat{k}(x)$, ${\hat w}_1(x)\text{ and }{\hat w}_2(x)$ are shown in the top left corners of the corresponding columns.
    }
    \label{fig: global regions inp}
\end{figure}

\begin{figure}[t]
    \centering
    {\hspace*{0cm}\small $\mid$------------------ Our -------------------$\mid$ ~~ im2im-uq ~~~ Conffusion}
    \includegraphics[width=\columnwidth,trim={4.2cm 3.3cm 3.5cm 3.3cm},clip]{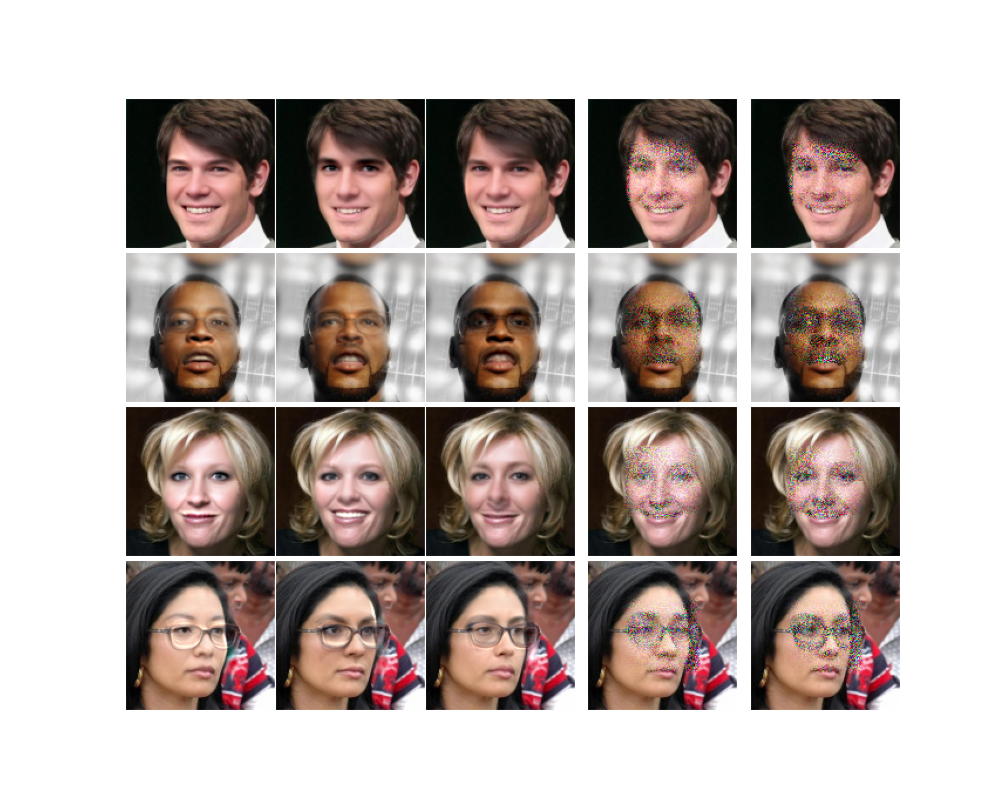}
    \caption{\textbf{Global Experiments:}  Images sampled uniformly from the estimated global uncertainty regions, referring to the inpainting task. Using DA-PUQ results with high-perceptual images, while im2im-uq \cite{angelopoulos2022image} and Conffusion \cite{horwitz2022conffusion} produce unlikely images. These results indicate that our uncertainty regions are significantly more confined than those of previous works.}
    \label{fig: global samples inp}
\end{figure}

\noindent In Section~\ref{sec: results} we presented global studies of our DA-PUQ and RDA-PUQ, focusing on their deployment in the colorization task.
Here, we extend this analysis by presenting additional global studies for super-resolution and inpainting, ensuring a more comprehensive assessment of our methods.

It is worth mentioning that the tasks of super-resolution and inpainting differ in nature from colorization.
In super-resolution and inpainting, the decay in the associated singular values of each posterior distribution occurs relatively slowly, indicating a more localized impact. This contrasts with the colorization task, where the decay in singular values is more rapid and pronounced, implying stronger pixel correlations.
Consequently, constructing global representations of uncertainty regions in the colorization task is effective, with strong guarantees involving small reconstruction errors over a large number of pixels using far fewer axes.

Nevertheless, we have applied our DA-PUQ and RDA-PUQ globally to the tasks of super-resolution and inpainting, and the quantitative results are depicted in Figure~\ref{fig: global results sr inp}. In both studies, we utilized 1500 samples for the calibration data and 500 samples for the test data. This is different from the global colorization study, where we used 1000 samples for both calibration and test data. This adjustment aims to narrow the gap between the true risks of unseen data and the concentration bounds employed in the calibration scheme, ultimately allowing us to provide more robust guarantees, including small coverage and reconstruction risks with high probability.

Additionally, we set $\alpha=\beta=\delta=0.1$ in both studies. However, in the super-resolution study, we maintained $q=0.95$, which is consistent with the setting used in the global colorization. In contrast, for inpainting, we chose $q=0.8$, indicating a softer reconstruction guarantee applicable to $80\%$ of the pixels within the missing window.

The results depicted in Figure~\ref{fig: global results sr inp} reveal that our method consistently yields significantly smaller uncertainty volumes compared to our local results presented in Section~\ref{sec: results} and previous research. However, this reduction in uncertainty volume comes at the cost of introducing a reconstruction risk, reaching a maximum of $\beta = 0.1$, which applies to $95\%$ of the pixels in super-resolution and $80\%$ of the pixels in inpainting.

Observe the improvement in interpretability that our DA-PUQ method brings to the table. Notably, the uncertainty regions generated by DA-PUQ consist of only $\sim$10 PCs within the full-dimensional space of the images. In contrast, the uncertainty regions produced by our RDA-PUQ experiments comprise $\sim$100 PCs, indicating a slower decay in the singular values of the posterior distribution associated with each uncertainty region.

Figures~\ref{fig: global regions sr} and ~\ref{fig: global regions inp} showcase selected uncertainty regions provided by our proposed DA-PUQ when applied globally to the super-resolution and inpainting tasks, respectively. Notably, the projected ground-truth images using only ${\hat k}(x)$ PCs resemble the originals. This observation indicates that the uncertainty region effectively captures the spread and variability among solutions while maintaining satisfying reconstruction errors.

In the inpainting task presented in Figure~\ref{fig: global regions inp}, the first two axes of our uncertainty regions exhibit semantic content, an indicator to our method's ability to consider spatial pixel correlation. The PCs effectively capture features such as sunglasses, eyebrows, and forehead, highlighting the unique strength of our approach in terms of interpretability. However, in the super-resolution task depicted in Figure~\ref{fig: global regions sr}, localized PCs emerge, implying that only a few pixel values are affected in each axis of uncertainty.

In Figures~\ref{fig: global samples sr} and \ref{fig: global samples inp}, we visually compare samples generated from the corresponding estimated uncertainty regions. These samples are obtained by uniformly sampling from the respective hyper-rectangle, then transforming to the image domain.
\Review{For further details regarding this study, please refer to Appendix~\ref{app: samples from uncertainty region}.}
This visual comparison\footnote{These results are better seen by zooming in, and especially so for the super-resolution task.} shows that samples extracted from our uncertainty regions exhibit higher perceptual quality compared to those generated by im2im-uq \cite{angelopoulos2022image} and Conffusion \cite{horwitz2022conffusion}. This observation implies that our method provides tighter uncertainty regions, whereas previous work results in exaggerated uncertainty regions that contain improbable images.

\subsection{Ablation Study}
\label{app: ablation study}

\begin{figure*}[t]
    \centering
    \includegraphics[width=\textwidth,trim={6.5cm 0 11cm 0},clip]{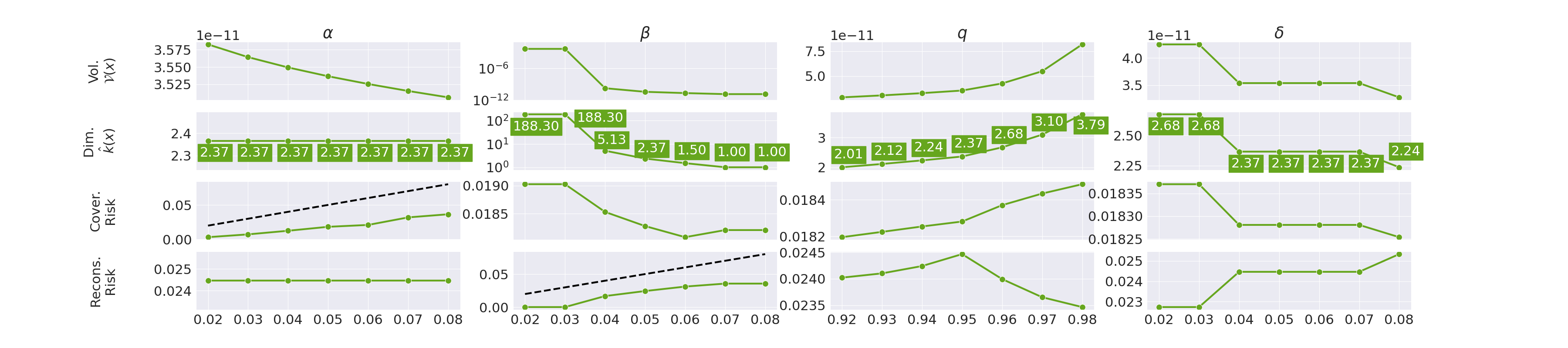}
    \caption{An ablation study of DA-PUQ in a locally applied colorization task on 8x8 RGB patch resolution. We examine the user-defined parameters $\alpha$, $\beta$, $q$, and $\delta$, showcasing their impact on the mean uncertainty volume, mean dimensionality, coverage risk, and reconstruction risk. The default values  are $\alpha=0.05$, $\beta=0.05$, $q=0.95$, and $\delta=0.05$ using $K=200$ PCs. Our results are depicted in green, with threshold values for guarantees highlighted in dashed black.}
    \label{fig: ablation}
\end{figure*}

\noindent We turn to introduce an ablation study on the user-specified parameters: $\alpha$, $\beta$, $q$, and $\delta$.
These parameters are used in the context of the statistical guarantees provided by our proposed method, and our objective is to offer a comprehensive understanding of how to select these parameters and their resulting impact on performance. To elaborate, $\alpha \in (0,1)$ is employed to ensure coverage, as indicated in Equation~\eqref{eq: theoretical coverage gaurentee}, while both $\beta \in (0,1)$ and $q \in (0,1)$ play a role in establishing the reconstruction guarantee, as defined in Equation~\eqref{eq: theoretical reconstruction gaurentee}. Additionally, the parameter $\delta \in (0,1)$ is used for controlling the error rate associated with both guarantees over the calibration data.

An effective calibration process relies on these user-specified parameters, $\alpha$, $\beta$, and $\delta$ approaching values close to zero, while $q$ should ideally approach 1. The choice of these parameters is guided by the amount of available calibration data.
In cases where a substantial calibration dataset is accessible, it becomes feasible to establish robust statistical assessments. This is manifested by the ability to employ smaller values for $\alpha$, $\beta$, and $\delta$, while favoring a higher value for $q$. For instance, achieving a $90\%$ coverage rate ($\alpha=0.1$), with a reconstruction error threshold of $5\%$ ($\beta=0.05$) across $95\%$ of the image pixels ($q=0.95$) serves as an illustrative example of such robust assessments.

It is worth noting that our primary aim in this work is to enhance the interpretability of the uncertainty assessment within the context of the inverse problems.
This is achieved through the methods we propose, DA-PUQ and RDA-PUQ.
Consequently, we strive to provide the user with a more concise set of uncertainty axes, referred to as the selected axes denoted as $\hat{B}(x) = \{ \hat{v}_1(x), \hat{v}_2(x), \dots, \hat{v}_{\hat{k}(x)}(x) \}$.
Our approach for selecting the reconstruction guarantee is geared towards a balance between precision and interpretability.
On one hand, we aim to establish a robust and stringent reconstruction guarantee to accurately capture the uncertainty of the posterior distribution across the $d$ dimensions.
On the other hand, we aim to incorporate a softer reconstruction guarantee that results in providing fewer axes of uncertainty thus enhancing interpretability.

Figure~\ref{fig: ablation} illustrates the quantitative results of the ablation study conducted on DA-PUQ, where we investigate the influence of the user-defined parameters, $\alpha$, $\beta$, $q$, and $\delta$, on DA-PUQ's performance.
It is important to note that the default settings in each study are the following: $\alpha=0.05$, $\beta=0.05$, $q=0.95$, and $\delta=0.05$, representing a spectrum of strengthening and softening parameter choices.

Analyzing the results, we observe that $\alpha$ primarily controls the coverage aspect. As $\alpha$ increases, the uncertainty intervals become narrower, leading to more tightly constrained uncertainty regions. This trend is evident in the reduction of the uncertainty volume metric. However, it is noteworthy that $\alpha$ has no impact on the reconstruction error, as the dimension and reconstruction risk remain relatively consistent across different choices of $\alpha$.

The parameter $\beta$ influences the reconstruction error, with even slight alterations affecting the number of selected axes, denoted as $\hat{k}(x)$.
On the other hand, the parameter $q$ has a relatively minor effect on performance. Adjusting $q$ does impact the uncertainty volume, with smaller dimensions resulting in a reduction in uncertainty volume. Higher values for $q$ lead to the selection of more PCs for the uncertainty assessment, involving more pixels in the reconstruction guarantee.

Referring to the parameter $\delta$, we observe minor changes in the coverage risk, while the reconstruction risk undergoes more significant changes. This suggests that errors in the uncertainty assessments tend to be more focused on the reconstruction guarantee rather than the coverage guarantee.

In terms of the precision and interpretability trade-off, the ideal scenario would involve selecting the smallest possible value for $\beta$, as demonstrated in Figure~\ref{fig: ablation} with $\beta = 0.02$. However, such a stringent guarantee would require the use of approximately $188.3$ PCs, which can harm interpretability. In this case, a softer guarantee, such as $\beta=0.04$, results in the use of only around $5.13$ PCs, striking a more balanced trade-off between precision and interpretability.

\subsection{Precision and Complexity Trade-off}
\label{app: tradeoff}

\begin{figure}[t]
    \centering
    \includegraphics[width=0.8\columnwidth,trim={1cm 1cm 0cm 1cm},clip]{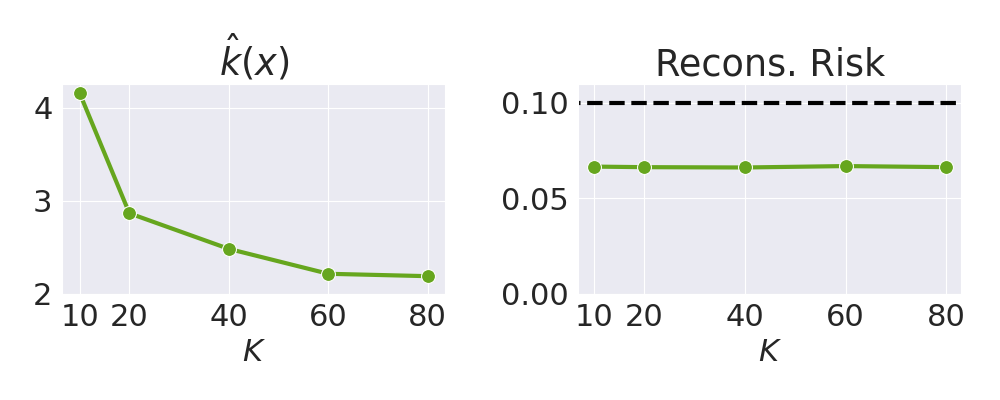}
    \caption{An analysis illustrating the precision-complexity trade-off of global DA-PUQ in the colorization task. The complexity aspect is presented by varying values of $K$, while precision is represented by the mean number of PCs provided for the user, denoted as $\hat{k}(x)$.
    The parameters setting is: $\alpha=0.1$, $\beta=0.1$, $q=0.95$ and $\delta=0.1$.
    Smaller $\hat{k}(x)$ values correspond to more accurate PCs, while lower values of $K$ indicate improved method complexity.}
    \label{fig: tradeoff}
\end{figure}

\noindent We now discuss the trade-off between precision and complexity in our work.
Precision here stands for the ability to accurately capture uncertainty within the posterior distribution across the $d$ dimensions, as reflected by the reconstruction risk.
Conversely, complexity involves two key aspects: the complexity associated with our diffusion model for generating posterior samples and the computational demands of PCA.
Both of these aspects are influenced by the chosen value of $K \leq d$, which serves both as the number of drawn samples and the overall number of initial PC's to work with. 
As for the complexity: 
(i) Assuming that a single diffusion iteration can be achieved in constant time, the complexity of generating $K$ samples is given by $O(IK)$, where $I\in\dN$ denotes the number of iterations in the diffusion algorithm; and (ii) For the PCA, the complexity is provided by $O(d^2K+d^3)$, where $K$ refers to the number of PCs.

Therefore, the value of $K \leq d$ plays a pivotal role in governing the precision-complexity trade-off across all our proposed methods: E-PUQ, DA-PUQ, and RDA-PUQ, all of which involve sampling and PCA.
The greater the number of PCs employed, the more precise our uncertainty assessment, at the expense of computational complexity, as discussed above.

In the case of E-PUQ, we achieve a complete uncertainty assessment at the computation cost of $K=d$. This leads to the effective reduction of the reconstruction risk to zero for all image pixels.
However, it's essential to recognize that in practical scenarios, such as the global applications illustrated in our empirical study in Section~\ref{sec: results} and in Appendix~\ref{app: additional global experiments}, conducting sampling and PCA with $K=d$ on high-dimensional data, such as $d=3\times128\times128$, becomes unfeasible.

Hence, we introduced DA-PUQ to enhance the method's computational efficiency by allowing $K \ll d$, thereby mitigating complexity.
To further enhance interpretability, we introduced $\hat{k}(x)$ in Equation~\eqref{eq: adaptive k}, which aims to reduce the number of PCs to be used (out of the already constructed $K$ PCs), while ensuring that the reconstruction guarantee is maintained with as few PCs as possible. This balance is demonstrated in Figure~\ref{fig: tradeoff}, where various values of $K$ showcase that the reconstruction risk remains unaffected, yet more uncertainty axes, $\hat{k}(x)$, are needed to uphold this equilibrium.

Given the challenge of determining an appropriate value for $K$ that ensures robust statistical guarantees, we introduced RDA-PUQ. This variant tunes $K$ to the lower value that fulfills the necessary statistical guarantees.

In Figure~\ref{fig: tradeoff}, we visually depict the precision-complexity trade-off through experiments involving different values of $K$ in the context of DA-PUQ's global application in colorization. Here, we illustrate the complexity of our method through the selection of varying $K$ values, where higher values imply higher complexity, as they require the generation of more samples and the construction of more PCs. Meanwhile, precision is assessed by examining the resulting $\hat{k}(x)$ values, where higher $\hat{k}(x)$ values correspond to situations where the uncertainty assessment is less accurate, signifying a higher reconstruction risk when employing all the $K$ PCs. Consequently, more axes are needed to maintain a balanced risk.

\subsection{Lower and Upper Corners}
\label{app: corners}

\begin{figure*}[t]
    \centering
    {\hspace{-4.8cm}\small $x$ ~~~~~~~~ $y$ ~~~~~~~~ $\hat{\mu}(x)$ ~~~~~~~ lo ~~~~~~~~ up ~~~~~~~~~ lo ~~~~~~~~ up ~~~~~~~~~ lo ~~~~~~~~ up}
    \includegraphics[width=\textwidth,trim={4cm 2cm 1cm 2.3cm},clip]{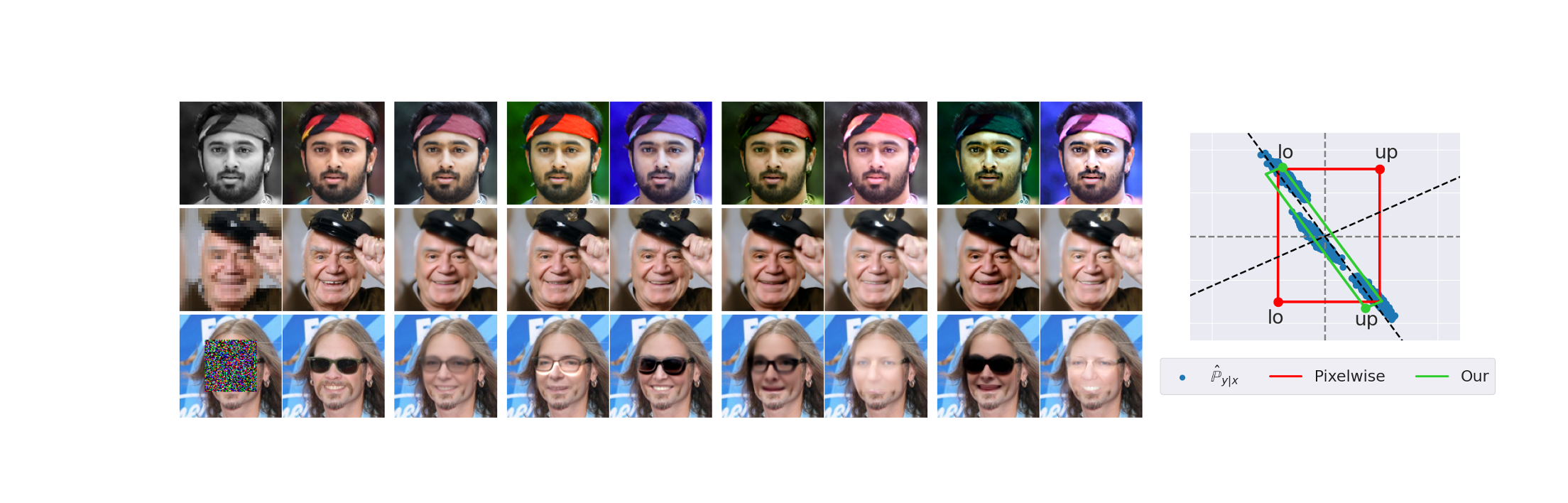}
    {\small $\mid$--------- Our ---------$\mid$ $\mid$----- im2im-uq ------$\mid$ $\mid$----- Conffusion -----$\mid$~~~~~}
    \caption{Visual analysis (left) of the lower and upper corners generated by our global DA-PUQ across three tasks: image colorization, super-resolution, and inpainting. On the right side, a 2D example illustrates an uncertainty region constructed by our approach in contrast to one produced by the pixelwise approach, demonstrating the distinction between the lower and upper corners in each approach.}
    \label{fig: corners}
\end{figure*}

\begin{figure*}[t]
    \centering
    {\hspace{0cm}\small $t=0.0$ ~~~~ $t=0.1$ ~~~~ $t=0.2$ ~~~~ $t=0.3$ ~~~~ $t=0.4$ ~~~~ $t=0.5$ ~~~~ $t=0.6$ ~~~~ $t=0.7$ ~~~~ $t=0.8$ ~~~~ $t=0.9$ ~~~~ $t=1.0$}
    \includegraphics[width=\textwidth,trim={9cm 2.3cm 7cm 2.3cm},clip]{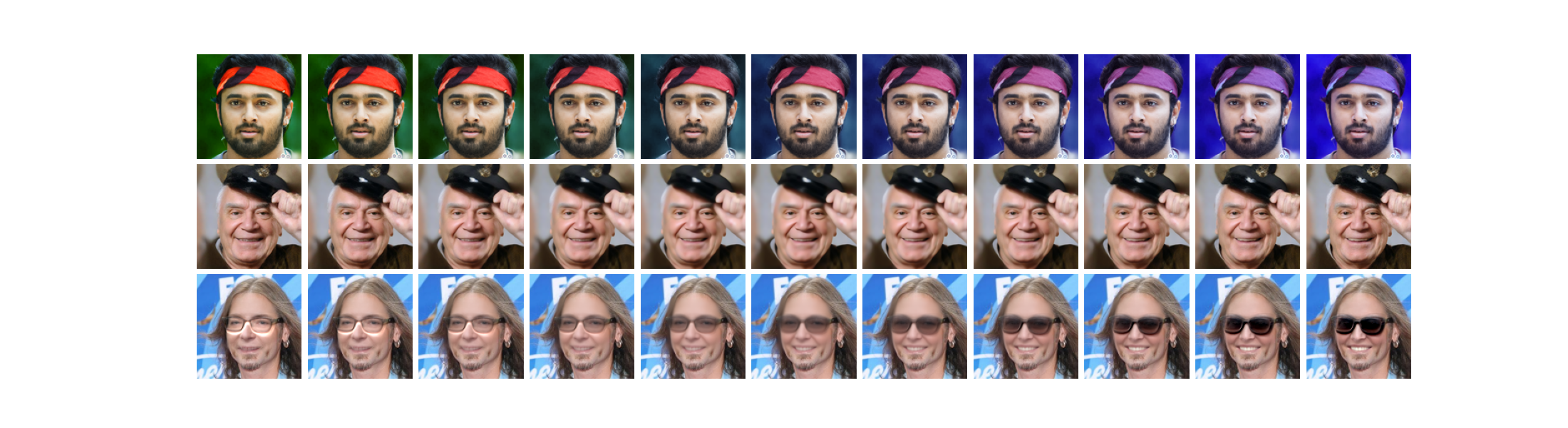}
    \caption{Visualization of the main ``boulevard'' within the uncertainty regions of DA-PUQ applied globally across three tasks: image colorization, super-resolution, and inpainting. The traversal along this path is obtained by a convex combination of the lower and upper corners, given by: $(1 - t) \cdot \text{lo}(x) + t \cdot \text{up}(x)$, where $t\in[0,1]$.}
    \label{fig: traverse}
\end{figure*}

\noindent We conclude this paper by providing a visual comparative analysis of lower and upper corners within uncertainty regions applied globally across the three tasks: image colorization, super-resolution, and inpainting.

Formally, the lower and upper corners within the image domain of an uncertainty region, are constructed using the intervals outlined in Equation~\eqref{eq: intervals}. These are defined as the following expressions:
the lower corner is defined as $\hat{\mu}(x) - \hat{V}(x) \hat{l}(x)$, and the upper corner is defined as $\hat{\mu}(x) + \hat{V}(x) \hat{u}(x)$.
Here, $\hat{V}(x)$ is a matrix comprising of the $K$ selected PCs from $\hat{B}(x)$ as it's columns, and $\hat{l}(x),\hat{u}(x)$ are column vectors of length $K$.

For example, when choosing to work within the pixel domain by selecting the standard basis, $\hat{B}(x) = {e_1, e_2 \dots e_d}$, where $e_i\in\dR^d$ represents the one-hot vector with a value of 1 in the $i^{th}$ entry, the lower and upper corners align with the lower and upper bounds presented in prior work \cite{angelopoulos2022image, horwitz2022conffusion} that operates in the pixel domain.

It is essential to note that in our work, we use the term \emph{corners} to emphasize that the lower and upper corners in the image domain do not establish intervals. This is in contrast to the pixelwise approach, which constructs intervals around each pixel, making the terminology ``lower and upper bound images'' more conceptually suitable.

In Figure~\ref{fig: corners} (right), we illustrate the difference between the lower and upper corners of our uncertainty region (depicted as green dots) and the lower and upper bounds of the pixelwise approach (depicted as red dots). This comparison is presented through a 2D example, demonstrating the process of constructing an uncertainty region for a posterior distribution using our method, in contrast to the pixelwise approach.

In Figure~\ref{fig: corners} (left), we provide a visual comparison between the lower and upper corners generated by DA-PUQ and the lower and upper bounds produced by \cite{angelopoulos2022image, horwitz2022conffusion}. It is evident that our lower and upper corners exhibit a higher perceptual quality compared to the lower and upper bounds from earlier pixel domain approaches. This suggests that the lower and upper corners represent more probable samples than those generated by the pixelwise approach.
Therefore, the uncertainty regions constructed by our approach are more confined compared to those constructed using the pixelwise approach. 

Interestingly, by traversing between the two corners of DA-PUQ by their convex combination, we essentially walk in the main ``boulevard'' of the uncertainty region. Figure~~\ref{fig: traverse} shows the resulting images in this path for the three applications considered: image colorization, super-resolution and inpainting.

\end{document}